%% file: main.tex
\documentclass{article}
\input{math_commands.tex}

\usepackage{microtype}
\usepackage{graphicx}
\usepackage{subcaption}
\usepackage{booktabs} 

\usepackage{hyperref}
\usepackage{url}
\usepackage{booktabs}       
\usepackage{amsfonts}       
\usepackage{nicefrac}       
\usepackage[titlenumbered,ruled]{algorithm2e}
\usepackage{amsmath}
\usepackage{algorithmic}
\usepackage{multirow}
\usepackage{multicol}
\usepackage{makecell}
\usepackage{subcaption}
\usepackage{eqparbox}
\usepackage{pifont}
\usepackage{xcolor}
\usepackage{footnote}
\usepackage{threeparttable} 
\usepackage{tabularx}
\newcommand{\mr}[2]{\multirow{#1}{*}{\begin{tabular}[c]{@{}c@{}}#2\end{tabular}}}

\usepackage{hyperref}

\usepackage[accepted]{icml2021}

\icmltitlerunning{Q-Rater: Non-Convex Optimization for Post-Training Uniform Quantization}

\begin{document}

\twocolumn[
\icmltitle{Q-Rater: Non-Convex Optimization for Post-Training Uniform Quantization}
\icmlsetsymbol{equal}{*}

\begin{icmlauthorlist}
\icmlauthor{Byeongwook Kim}{equal,samsung}
\icmlauthor{Dongsoo Lee}{equal,samsung}
\icmlauthor{Yeonju Ro}{samsung}
\icmlauthor{Yongkweon Jeon}{samsung}
\icmlauthor{Se Jung Kwon}{samsung}
\icmlauthor{Baeseong Park}{samsung}
\icmlauthor{Daehwan Oh}{samsung}
\end{icmlauthorlist}

\icmlaffiliation{samsung}{Samsung Research, Seoul, Republic of Korea}

\icmlcorrespondingauthor{Byeongwook Kim}{byeonguk.kim@samsung.com }

\vskip 0.3in
]

\printAffiliationsAndNotice{\icmlEqualContribution} 

\newcommand{\Nin}{N_{in}}
\newcommand{\Nout}{N_{out}}
\newcommand{\Ns}{N_{s}}
\begin{abstract}
Various post-training uniform quantization methods have usually been studied based on convex optimization.
As a result, most previous ones rely on the quantization error minimization and/or quadratic approximations.
Such approaches are computationally efficient and reasonable when a large number of quantization bits are employed.
When the number of quantization bits is relatively low, however, non-convex optimization is unavoidable to improve model accuracy.
In this paper, we propose a new post-training uniform quantization technique considering non-convexity.
We empirically show that hyper-parameters for clipping and rounding of weights and activations can be explored by monitoring task loss.
Then, an optimally searched set of hyper-parameters is frozen to proceed to the next layer such that an incremental non-convex optimization is enabled for post-training quantization.
Throughout extensive experimental results using various models, our proposed technique presents higher model accuracy, especially for a low-bit quantization.

\end{abstract}

\section{Introduction}

The model size of deep neural networks (DNNs) is rapidly growing to support various complex target applications with increasing target accuracy goals.
Hence, numerous model compression techniques are being actively studied to enable DNN inference operations for a given service response time while computing resources are limited.
Such compression techniques include parameter pruning \cite{DNS, SHan_2015}, low-rank approximation \cite{SVD2013, SVD_Projection}, knowledge distillation \cite{distillation, distillation2018}, and quantization \cite{binaryconnect, Hubara2016}.
In this paper, we discuss a quantization method specifically designed to preserve the model accuracy of DNNs.

Since quantization plays a major role to determine not only the (expensive) off-chip memory bandwidth but also the basic design principles of core arithmetic units performing DNN operations \cite{Hubara2016, lin2016fixed}, researchers are paying a lot of attention to the advance of DNN-dedicated quantization methods.
For example, the Straight-Through Estimator (STE) \cite{binaryconnect} is widely used for quantization-aware training to enable binary neural networks in which each weight and activation can be represented by one bit.
In order to improve model accuracy after quantization, hyper-parameters for the quantization format can be designed to be differentiable and trainable \cite{sait_uniform, adaptive, ternary2017}.
Recently, even encryption techniques and training algorithms are introduced to implement sub-1-bit quantization with negligible accuracy drop \cite{flexor}.
While quantization-aware training is effective to improve model accuracy, a wide range of post-training quantization schemes are also being considered because 1) DNN designers may not have enough expertise to consider model compression, and 2) model compression engineers may not be able to access the whole training dataset.
As a result, numerous sophisticated post-training quantization algorithms are introduced \cite{bit-split, OCS}, and are being served by various DNN model development tools \cite{tensorflow2015-whitepaper, pytorch}.

\begin{figure}
\vskip 0.2in
    \begin{center}
    \includegraphics[width=0.9\linewidth]{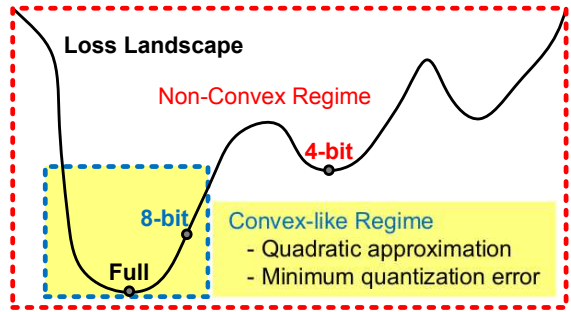}
    \end{center}
    \caption{A loss landscape example where 8-bit quantization can be a convex-like optimization while optimal 4-bit quantization needs to be achieved by a non-convex optimization.}
    \label{fig:loss_landscape}
    \vskip -0.2in
\end{figure}

Most existing post-training quantization algorithms rely on convex-like optimization, essentially because fine-tuning or retraining as non-convex optimization is not available.
For example, quantization errors on weights \cite{Greedy_Quan, OCS, balanced} or layer outputs \cite{adaround, facebook_quan, bit-split} are mainly minimized.
When such minimization is NP-hard problems, quadratic approximations can be adopted to simplify the minimization process \cite{limitquant2017, adaround}.
Note that when the amount of weight perturbation through quantization is large, then minimizing the quantization error may not be a convex-like optimization as shown in Figure~\ref{fig:loss_landscape}.
In addition, if a given loss surface of a pre-trained model is not smooth enough \cite{loss_surface}, then even post-training 8-bit quantization can be translated into a non-convex optimization problem.
As a comprehensive post-training quantization, thus, an approach considering non-convexity is necessary.

Intuitively, it would be challenging to study a thorough non-convex analysis for post-training quantization.
In this paper, based on the recognition that post-training inherently requires non-convex approaches, we propose a new post-training quantization method, called Q-Rater, that is especially useful for low-bit quantization.
Instead of minimizing quantization errors, Q-Rater searches (not computes) hyper-parameters for quantization to minimize the task loss.
The contributions in this paper can be summarized as follows:
\begin{itemize}
    \item We present some examples suggesting that minimizing quantization error may not be tightly correlated with minimizing a training loss function.
    \item We propose new methods to find hyper-parameters determining clipping threshold values and rounding schemes. Such hyper-parameters are obtained by evaluating training loss function through a grid search for a layer. Then, searched hyper-parameters can be fixed for the next layer exploration.
    \item We show that bias correction needs to be selectively performed per layer.
    \item Experimental results describe that Q-Rater yields higher model accuracy compared to previous techniques relying on convex-like optimizations, especially when the number of quantization bits is low.
\end{itemize}

\section{Weight quantization strategy}

\begin{figure}
\vskip 0.2in
    \begin{center}
    \includegraphics[width=0.9\linewidth]{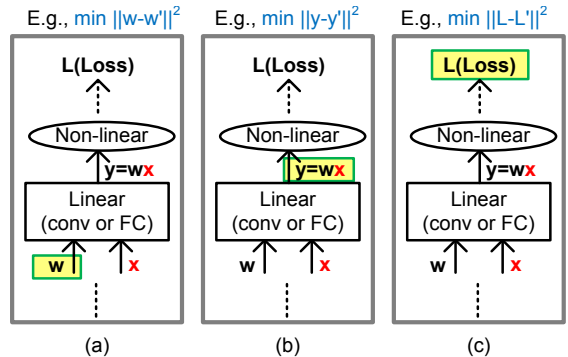}
    \end{center}
    \caption{Comparison on three post-training quantization strategies. Quantization algorithm can be designed to minimize (a) quantization error on weights, (b) reconstruction error by quantization, or (c) task loss error by quantization.}
    \label{fig:quant_strategy}
    \vskip -0.2in
\end{figure}

Let $\mW$ and $\vx$ represent weights and inputs of a layer, respectively.
By large, post-training weight quantization strategies can be categorized into three schemes as shown in Figure~\ref{fig:quant_strategy} depending on the selection of the target variable to be minimized.
Note that Figure~\ref{fig:quant_strategy}(a) and \ref{fig:quant_strategy}(b) represent convex optimizations to calculate quantized $\mW$.

\paragraph{Weight only}
As the simplest method in Figure~\ref{fig:quant_strategy}, weights can be quantized in a local optimization manner, i.e., weight quantization does not consider other layers nor activations.
Various objective functions using the difference between $\mW$ and quantized weights $\mW'$ have been suggested.
For example, the mean squared error (MSE) using a histogram \cite{shin2016MSE} or a predetermined distribution model \cite{ACIQ} can be minimized to obtain quantized weights.
Minimizing KL divergence between full-precision weight distribution and quantized weight distribution is also proposed \cite{tensorRT}.
Since input data is not utilized, the quantization process can be simple and fast \cite{DFQ} even though the correlation between weight quantization and task loss is not deeply investigated.

\paragraph{Layer output objective}
Given a specific domain of input data, quantized weights obtained only by using weights would not produce the best-quantized layer outputs \cite{facebook_quan}.
Correspondingly, to take into account the statistical properties of input domains, samples of inputs can be fed into the network and the quantization error on layer outputs can be minimized \cite{facebook_quan, bit-split}.
Suppose that $\mX$ is a set of input samples and the objective function is given as $\min ||\mW\mX - \mW'\mX'||^2$.
Then, compared to the case of Figure~\ref{fig:quant_strategy}(a), the computational complexity (to solve $\min ||\mW\mX - \mW'\mX'||^2$) may significantly increase because of large $\mX$.
For example, feature size can be larger than the weight size in convolutional neural networks (CNNs) \cite{resnet}.
In addition, the number of features (i.e., input sets) needs to be much larger than 1.
For instance, in the case of ResNet-101 on ImageNet, the number of elements in $\mX$ can be about 80M with 100 input samples (56x56 output size and 256 channels).

\paragraph{Task loss objective}
Weight quantization to produce the minimum task loss can be the most effective strategy if we find a sophisticated relationship between weight manipulation for quantization and corresponding task loss change.
Unfortunately, because the task loss function of DNNs is non-convex \cite{deeplearningbook}, there is no analytical solution to find the form of post-training quantized weights.
As a result, various approximations are being suggested mainly by using quadratic approximations that imply quantization is performed in a convex-like regime \cite{adaround, loss-aware}.
Note that such approximations may hold only for a large number of quantization bits as we demonstrate in Section 4.

\section{Activation quantization strategy}
Unlike weights that can be quantized and fixed in advance, activations should be quantized on the fly during inference.
In other words, weights are static data while activations are dynamic data that change during inference.
Hence, quantization techniques dedicated to static data cannot be adopted for activation quantization.
Accordingly, there are relatively fewer studies for activation (post-training) quantization compared to weight quantization.
Thus, hyper-parameters regarding activation quantization are usually drawn by a sampling of activations and estimating a distribution.
For example, the moving average of activation values obtained by feeding input samples can be a clipping threshold \cite{jacob2018quantization}.

\section{Issues on the previous methods}
Some major issues on previous techniques for post-training quantization include that 1) non-convex properties are not exploited; 2) discussions on activations are somewhat overlooked; and 3) consequently, the impact of quantization on task loss is loosely coupled with an objective function to be minimized.

\begin{figure}[t]
    \centering
    \vskip 0.2in
    \includegraphics[width=1.0\linewidth]{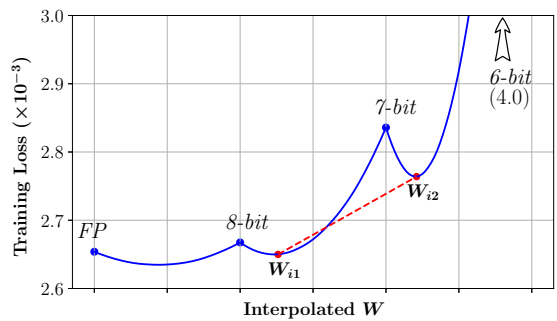}
    \caption{A simple 1-D trajectory investigation through interpolations of weight sets quantized from the same pre-trained ResNet-32 model on CIFAR-10.} 
    \label{fig:1d_trajectory}
    \vskip -0.2in
\end{figure}

Let us first show that quantization needs to be considered as a non-convex optimization.
When $\mW_{q1}$ and $\mW_{q2}$ are two quantized weights with different number of bits, interpolated $\mW_q$ is given as
\begin{equation}
    \mW_q = (1-\alpha)\mW_{q1} + \alpha \mW_{q2}, \; \mathrm{for} \; \alpha \in [0,1].
\end{equation}
Even though a thorough trajectory study between different parameter sets is highly complicated, evaluating a loss function $L(\mW_q)$ with sweeping $\alpha$ can provide a 1-D interpolated trajectory cross-section to seek a counter-example of convex-like regime \cite{1d_plot_trajectory}.
Figure~\ref{fig:1d_trajectory} traces the training loss function when weight sets are interpolated using weights of full-precision or quantized.
Note that if a function $f(x)$ is convex, then for any two points $x_1$ and $x_2$, we obtain
\begin{equation}\label{eq:convexity}
    f(\lambda x_1 + (1-\lambda )x_2) \le \lambda f(x_1) + (1-\lambda) f(x_2),
\end{equation}
when $\forall \lambda \in [0,1]$.
In Figure~\ref{fig:1d_trajectory}, we find that \textbf{training loss function is non-convex} because training loss value can be even lower when weight sets are interpolated such that Eq.~\ref{eq:convexity} does not hold (e.g., training loss function is non-convex between $\mW_{i1}$ and $\mW_{i2}$ in Figure~\ref{fig:1d_trajectory}).
Hence, Figure~\ref{fig:1d_trajectory} supports our claim that convex-like optimizations could miss better quantization schemes.

\begin{figure*}[t]
    \centering
    \vskip 0.2in
    \includegraphics[width=\linewidth]{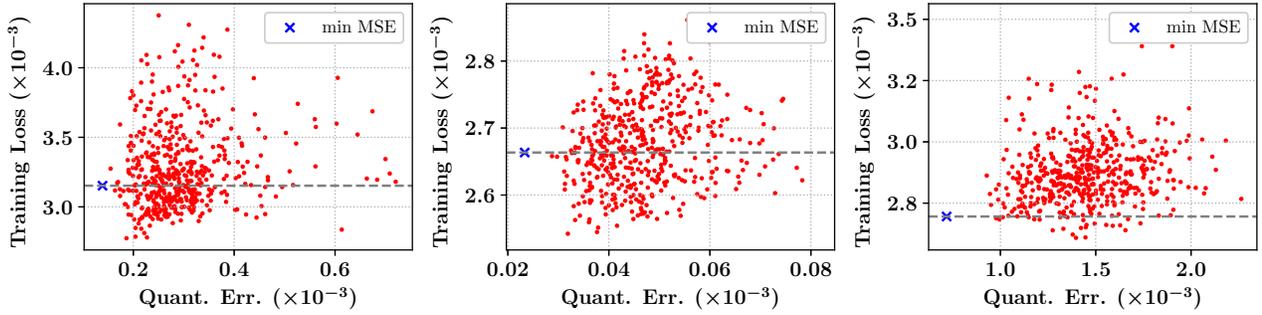}
    \caption{Correlation between quantization error and training loss of ResNet-32 (on CIFAR-10) when each weight of the 1st, the 16th, or the 31st layer is quantized by using 4 bits.}
    \label{fig:resnet32_qerr_and_loss_4bit}
    \vskip -0.2in
\end{figure*}

Non-convexity of quantization can also be confirmed by investigating the relationship between quantization error and training loss.
Suppose that $n$-bit quantization schemes lie in a convex-like regime.
If it is the case, increasing quantization error would result in increasing training loss with a high correlation.
As a case study, we apply 4-bit weight quantization to a few layers in ResNet-32 on CIFAR-10 using various quantization schemes that we discuss in the next section.
Figure~\ref{fig:resnet32_qerr_and_loss_4bit} illustrates that a correlation between quantization error (on weights) and training loss is not high such that low-bit quantization can be regarded as a non-convex problem.
Similar observations are also reported in \cite{adaround, facebook_quan} (where reconstruction error is minimized as a practical solution).

Analytical solutions based on the first two strategies in Figure~\ref{fig:quant_strategy} may not be appropriate for the post-training quantization of high non-convexity.
In this work, we show that there is a way to find quantization schemes to reduce model accuracy reduction directly without quantization error minimization on weights or layer outputs.



\section{Q-Rater: holistic non-convex quantization}

Unlike most previous methods proposing convex optimizations to obtain hyper-parameters for quantization, Q-Rater explores various hyper-parameters and evaluates corresponding training loss values using a grid search for a layer.
Then, Bayesian optimization is performed to fine-tune a set of hyper-parameters in a layer.
Once the fine-tuned hyper-parameters are achieved for a certain layer, then those hyper-parameters are frozen, and Q-Rater proceeds to the next target layer.
For Q-Rater, we consider rounding and clipping as the underlying quantization operations.
The bias correction method is also discussed in the context of non-convexity.
In this section, Q-Rater operations (i.e., rounding, clipping, and bias correction) are individually introduced and then combined to present the overall impact on model accuracy.

Throughout this paper, even though Q-Rater does not rely on particular quantization formats, we assume \textbf{layer-wise} and \textbf{symmetric} quantization structure for both weights and activations.
Such quantization structure is highly practical because 1) a floating-point scaling factor is shared by all elements in a layer, and thus, whole matrix multiplications (or convolutions) can be performed in fixed-point formats, and 2) computations for zero-point are not necessary \cite{jacob2018quantization, bit-split}.
We show that Q-Rater with such simple and computationally efficient quantization formats can maintain reasonable model accuracy for low-bit quantization.

\subsection{Rounding scheme of Q-Rater}
Assume that a (layer-wise) clipping threshold is given as $Th_c (>0)$.
As the first step of symmetric quantization, a weight is clipped to be $w_c = \max (\min(w, Th_c ),-Th_c)$.
Rounding is then performed to map a continuous value $w_c$ into one of the discrete values that are pre-determined for uniform quantization.
Note that rounding-to-nearest (RTN) has been dominating for uniform quantization since the per-weight difference becomes the smallest after mapping.
To be more specific, since we consider symmetric quantization, a scaling factor $s$ is given as $s=Th_c/(2^{q-1}-1)$ when $q$ is the number of quantization bits.
Then an integer $w_r$ can be obtained by RTN as $w_r = \left \lfloor w_c / s + 0.5 \right \rfloor$.
Training loss can be further reduced when weight is rounded up or down depending on the interaction between a perturbed weight and task loss.
Recently proposed adaptive rounding, AdaRound \cite{adaround}, investigates such an interaction using quadratic approximations and obtains rounding results by minimizing an asymmetric reconstruction formulation using gradient descent.
The rounding principles of AdaRound, however, cannot be applied to activations since rounding needs to be a fixed hyper-parameter for each weight.

To allow rounding up and down for activations as well, Q-Rater takes into account (unequal) ranges of a variable to be quantized.
In other words, Q-Rater divides a range of weights or activations into unequal parts where each part is mapped into the same discrete value.
Specifically, we propose the two following rounding schemes to achieve a quantized weight $w_q$ when $f_r(\cdot)$ decides the rounding (activations follow the same procedures).
\paragraph{1st-order rounding scheme ($\gamma_n$):}
\begin{equation}
    w_q = s \cdot \left \lfloor \frac{w_c}{s} + 0.5 + f_r(w_c, \gamma_n ) \right \rfloor,
    \label{eq:1st-order-eq1}
\end{equation}
\begin{equation}
    f_r(w_c, \gamma_n) = 0.5 \cdot \sign(w_c\gamma_n) \cdot |\gamma_n |^{|w_r|}, 
    \label{eq:1st-order-eq2}
\end{equation}
where $\gamma_n \in [-1,1]$ is a hyper-parameter.
If $\gamma_n=0$, Eq.~\ref{eq:1st-order-eq1} is equivalent to RTN.

\paragraph{2nd-order rounding scheme ($\gamma_n, \gamma_s$):}

\begin{equation}
    w_q = s \cdot \left \lfloor \frac{w_c}{s} + 0.5 + f_r(w_c, \gamma_n, \gamma_s ) \right \rfloor,
    \label{eq:2nd-order-eq1}
\end{equation}
\begin{equation}
\begin{split}
    f_r(w_c, \gamma_n, \gamma_s) = 0.5 \cdot \sign(w_c\gamma_n (\gamma_s \cdot 2^{q-1} - |w_r|)) \\
    \cdot |\gamma_n |^{|||w_r| - \gamma_s \cdot 2^{q-1}| - \beta|},
\end{split}
    \label{eq:2nd-order-eq2}
\end{equation}
where $\beta = 2^{q-2}$, and $\gamma_n \in [-1,1]$ and $\gamma_s \in [0,1]$ are hyper-parameters.
$\gamma_n=0$ produces RTN regardless of $\gamma_s$.

\begin{figure*}[t]
    \centering
    \vskip 0.2in
    \begin{subfigure}[b]{0.45\textwidth}
    \centering
    \includegraphics[width=\textwidth]{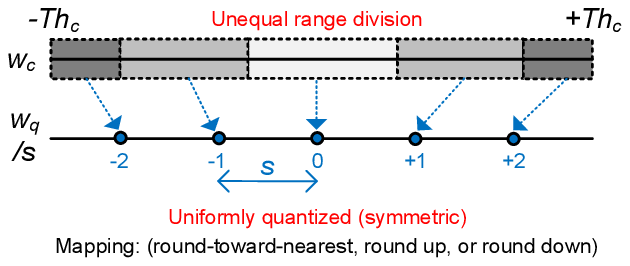}
    \caption{Proposed mapping method by unequal range division of a weight or activation.}
    \label{fig:rounding_left_figure}
    \end{subfigure}
    \hfill
    \begin{subfigure}[b]{0.48\textwidth}
    \centering
    \includegraphics[width=\textwidth]{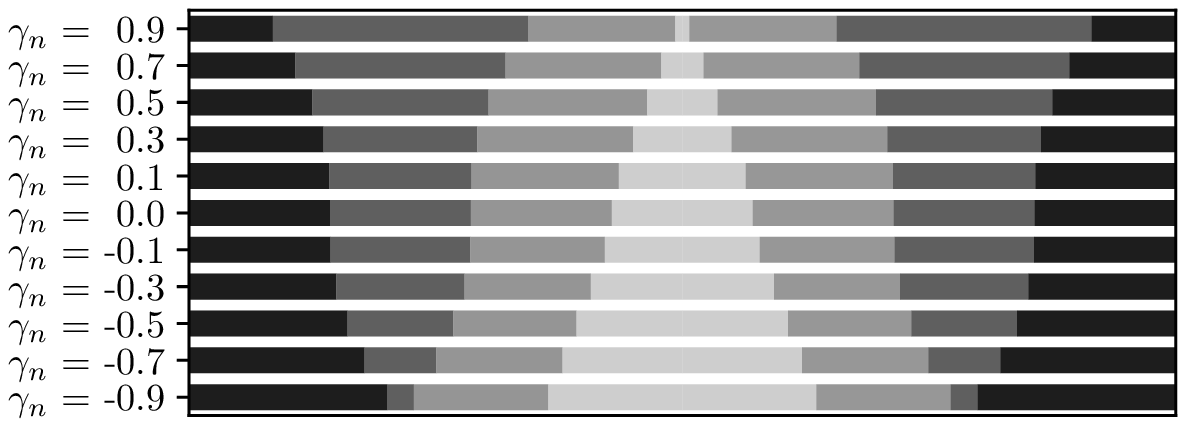}
    \caption{Examples of unequal range division by 2nd-order rounding scheme when $\gamma_s=0.5$ with various $\gamma_n$.}
    \label{fig:rounding_right_figure}
    \end{subfigure}
    \caption{Proposed rounding scheme based on unequally divided ranges to be mapped to quantized values.}
    \vskip -0.2in
    \label{fig:rounding_example}
\end{figure*}

\bigskip

The proposed rounding scheme is illustrated in Figure~\ref{fig:rounding_example} (due to the space limit, all illustrations about 1st-order rounding are provided in Appendix A).
Since $f_r(\cdot)$ has the range of $[-0.5,0.5]$ for both proposed 1st- and 2nd-order round schemes, Eq.~\ref{eq:1st-order-eq1} and \ref{eq:2nd-order-eq1} can select one of three rounding schemes, namely, RTN, round up, and round down.
$\gamma_n$ controls the amount of inequality among mapping ranges while $\gamma_s$ (for 2nd-order rounding) decides from where mapping ranges start to increase or decrease.


\begin{figure*}[t]
    \centering
    \vskip 0.2in
    \includegraphics[width=0.95\linewidth]{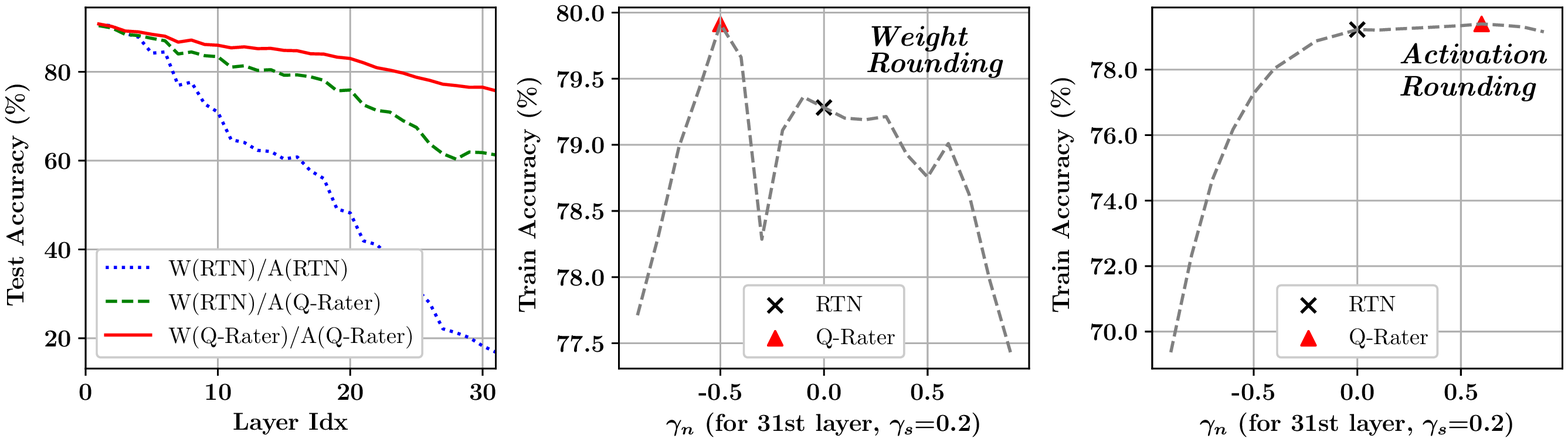} 
    \caption{Test accuracy comparison using ResNet-32 model (on CIFAR-10) when weights and/or activations are rounded by RTN or the proposed (2nd-order) rounding scheme that evaluates train accuracy with various $\gamma_n$ and $\gamma_s$. 3-bit quantization is performed from the second layer to the last one incrementally and the clipping method follows MSE minimization.}
    \label{fig:comparison_non_linear_2}
    \vskip -0.2in
\end{figure*}

Now let us explain how to find hyper-parameters for rounding.
First, for a target layer, we find a clipping threshold $Th_c$.
Then, we sweep $\gamma_n$ and $\gamma_s$, and find a particular set of $\gamma_n$ and $\gamma_s$ that correspond to the best training loss.
Once the optimal $\gamma_n$ and $\gamma_s$ are obtained through such a grid search, then we quantize the weights of a target layer.
Those quantized weights are fixed and we proceed to the next layer.

We apply different rounding methods to activations and/or weights of the ResNet-32 model on CIFAR-10 while quantization is performed incrementally from the second layer to the last one (note that the first layer is too small in ResNet-32).
To be focused on the impact of the new rounding scheme, we utilize MSE minimization to compute the clipping threshold $Th_c$ as verified to be effective in \cite{OCS} while we propose a new clipping technique in the next subsection.
Sweep interval is 0.1 for both $\gamma_n$ and $\gamma_s$.
Figure~\ref{fig:comparison_non_linear_2} compares quantization results of ResNet-32 model on CIFAR-10 using RTN or 2nd-order Q-Rater rounding scheme ($q=3$ for both weights and activations).
When $\gamma_s=0.2$, sweeping $\gamma_n$ to quantize the weights of the 31st layer yields high variations on train accuracy.
Note that for both weights and activations, RTN (i.e., $\gamma_n=0$) does not provide the best train accuracy and the optimal $\gamma_n$ is far from zero.
A set of $\gamma_n$ values optimized separately for each layer leads to significantly improved test accuracy as layers are quantized incrementally as shown on the left of Figure ~\ref{fig:comparison_non_linear_2}.
For the experimental results using ResNet-18 and MobileNetV2 on ImageNet (with only 20K input samples), refer to Appendix A. 
Throughout comprehensive experiments, the 2nd-order rounding scheme offers slightly better results than the 1st-order rounding scheme (shown in Appendix A).
For the remainder of this paper, thus, we choose the 2nd-order rounding scheme.

\subsection{Clipping method of Q-Rater}
\label{sec:clipping}

Most DNN models exhibit a bell-shaped distribution for weights and activations \cite{OCS}.
Hence, a few outliers in distribution can affect the overall quality of uniform quantization decisively.
Clipping method, therefore, is being widely used for post-training quantization \cite{ACIQ, jacob2018quantization, OCS} as a trade-off between the quantization resolution and the amount of outliers' quantization error.

\begin{figure*}[t]
    \centering
    \vskip 0.2in
    \includegraphics[width=0.95\linewidth]{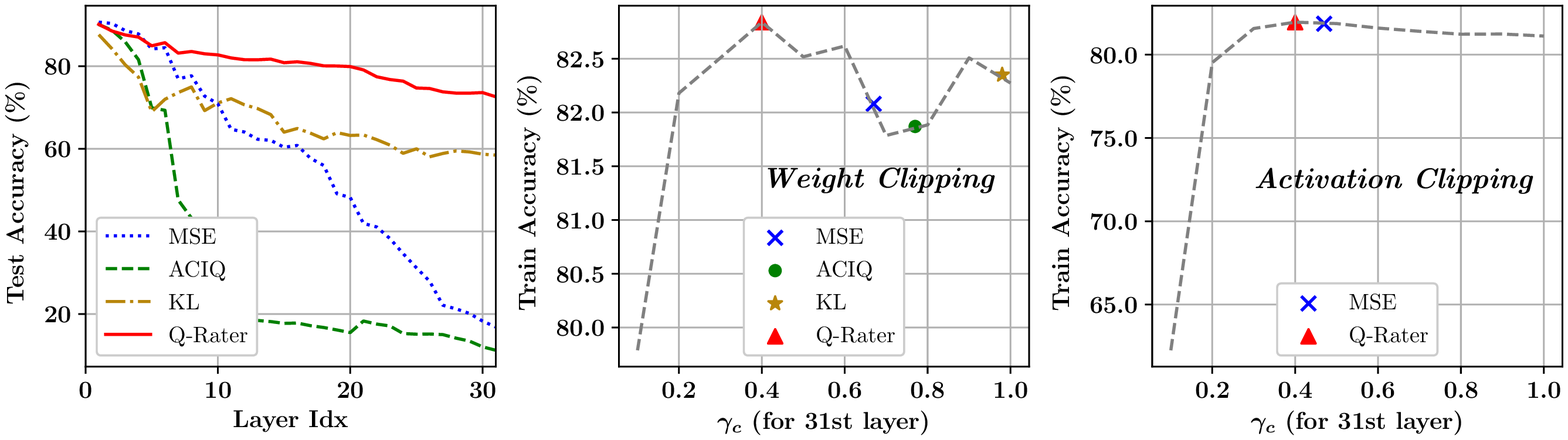}
    \caption{Test accuracy comparison using MSE, ACIQ, KL, or our proposed clipping method with RTN rounding scheme for weights and activations of ResNet-32 model on CIFAR-10 ($q=3$ for both weights and activations). For each layer, Q-Rater sweeps $\gamma_c$ and evaluates corresponding train accuracy.}
    \label{fig:clipping_exp}
    \vskip -0.2in
\end{figure*}

Recently proposed clipping methods include MSE minimization (i.e., L2-norm between the full-precision values and quantized values is minimized \cite{shin2016MSE, sung2016_mse_clip}), ACIQ (i.e., minimizing MSE between a pre-determined distribution model and the full-precision model \cite{ACIQ}), and KL divergence minimization \cite{tensorRT}.
An additional method to control outliers is outlier channel splitting (OCS) that duplicates channels containing outliers, and then halves activations or weights without modifying the functionality of a model \cite{OCS}.
OCS becomes more effective when combined with existing clipping methods \cite{OCS}.
Note that non-convexity is not recognized for those previous clipping techniques.

For Q-Rater, we introduce a hyper-parameter $\gamma_c \in (0.0, 1.0)$ to determine $Th_c = \max(\mW) \cdot \gamma_c$ when $\max(\mW)$ describes the maximum elements of a set of full-precision weights $\mW$ (clipping for activations follows the same structure).
Similar to our rounding schemes, we sweep $\gamma_c$ from 0 to 1 and investigate the corresponding training accuracy.
For experiments, after an optimal $\gamma_c$ (producing the best training accuracy) is obtained through a sweeping, RTN is followed for each weight and activation (RTN is used to study the impact of clipping method independently).
We perform an incremental quantization and Figure~\ref{fig:clipping_exp} presents the results using the ResNet-32 model on CIFAR-10 with different clipping methods.
It is clear that for low-bit quantization (e.g., $q=3$ in Figure~\ref{fig:clipping_exp}), conventional techniques may not produce the best training accuracy.
When such $\gamma_c$ optimization for a layer is conducted incrementally through all target layers, Q-Rater presents significantly higher test accuracy compared to MSE, ACIQ, and KL.
Experimental results using the ResNet-18 model on ImageNet are presented in Appendix B. 

\begin{figure}
    \centering
    \vskip 0.2in
    \includegraphics[width=0.8\linewidth]{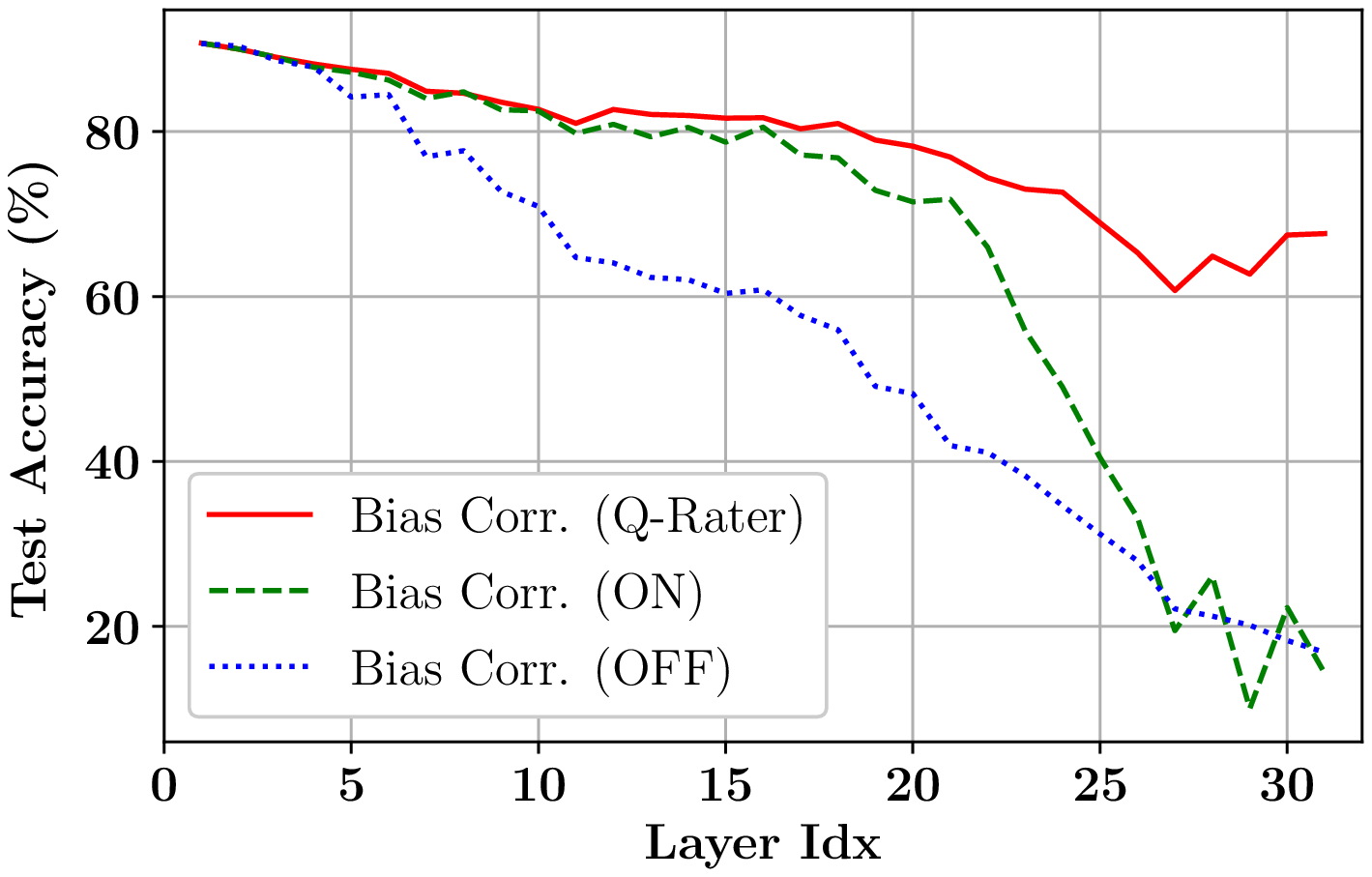}
    \caption{Test accuracy of ResNet-32 model on CIFAR-10 when bias correction is always applied, never applied, or selectively applied by Q-Rater. Weights and activations are quantized (by using $q=3$) incrementally with RTN and MSE clipping.}
    \label{fig:comparison_bias_corr}
    \vskip -0.2in
\end{figure}

\subsection{Bias correction of Q-Rater}
Bias correction is an operation to compensate for the biased error in output activations after quantization.
The amount of shift induced by quantization is diminished by adjusting the bias parameters of the neurons or channels because shifted output activations through quantization may degrade the quantization quality of the next layer \cite{fighting_bias, DFQ}.
The amount of shift can be calculated as the expected error on the output activations that can be expressed as
\begin{equation}
    \E[\vy] -\E[\vy'] = \E[\mW\mX] - \E[\mW'\mX'].
    \label{eq:bias_corr}
\end{equation}
Then, the expected error (or shift) of the output activations is subtracted from the corresponding layer's bias terms.

\begin{table*}[t]
        \small
        \caption{Top-1 test accuracy (\%) of various models quantized by previous methods or by Q-Rater. Model names are annotated with dataset and test accuracy of full-precision. All results are accompanied by \textbf{layer-wise} and \textbf{symmetric} quantization.}
        \vskip 0.15in
        \label{table:ablation_study_final_results}
        \centering
        \begin{threeparttable}
        \begin{tabular}{cc|ccc|ccccc}
        \hline
        \mr{2}{Model}                         & \mr{2}{W/A\\bits}          & \multicolumn{3}{c|}{Clip (+RTN)\tnote{1}}  & \multicolumn{5}{c}{Q-Rater (Proposed)} \\ \cline{3-10} 
                                              &                      & MSE   & ACIQ  & KL    & \ding{202}(Round)\tnote{2} & \ding{203}(Clip) & \ding{204}(Bias) & \ding{202}+\ding{203}  & \ding{202}+\ding{203}+\ding{204} \\ \hline
        \mr{3}{\textbf{ResNet-32}\\\textit{on CIFAR-10}\\(92.63)}  
                & 5/5                  & 90.64 & 75.68 & 90.45      & 91.70      & 91.68      & 91.64   & 91.93   & 91.74 \\
                & 4/4                  & 84.46 & 24.57 & 71.32      & 89.03      & 89.60      & 88.47   & 89.71   & 90.24 \\
                & 3/3                  & 16.86 & 11.23 & 58.48      & 75.74      & 72.61      & 67.63   & 77.35   & 79.69 \\ \hline
        \mr{3}{\textbf{ResNet-18}\\\textit{on ImageNet}\\(69.69)}
                & 8/8                  & 69.25 & 68.15 & 68.39      & 69.27      & 69.35      & 69.50   & 69.33   & 69.43 \\
                & 6/6                  & 66.75 & 59.78 & 65.64      & 67.34      & 67.91      & 67.42   & 67.95   & 68.46 \\
                & 4/4                  & 32.16 &  2.10 & 11.27      & 45.18      & 52.59      & 51.59   & 54.87   & 59.17 \\ \hline
        \mr{3}{\textbf{MobileNetV2}\\\textit{on ImageNet}\\(71.78)}         
                & 8/8                  & 69.81 & 69.98 & 69.98      & 71.01      & 70.73      & 71.34   & 70.92   & 71.25 \\
                & 6/6                  & 36.82 & 11.08 & 38.63      & 64.15      & 61.04      & 64.89   & 65.33   & 67.64 \\
                & 4/4                  & 0.12  &  0.39 &  0.22      & 7.08       &  4.67      &  6.58   & 15.58   & 25.35 \\ \hline
        \end{tabular}
        \begin{tablenotes}
            \footnotesize
            \item[1] Reference code: https://github.com/cornell-zhang/dnn-quant-ocs
            \item[2] 2nd-order rounding scheme is applied.
        \end{tablenotes}
    \end{threeparttable}
    \vskip -0.1in
\end{table*}

Bias correction has been a supplementary and optional technique for quantization.
For example, bias correction is not introduced in \cite{OCS} while it is playing a key role in enhancing model accuracy in \cite{fighting_bias, DFQ}.
In the context of non-convexity, Q-Rater compares two model accuracy values evaluated with or without bias correction for a layer.
Only when bias correction turns out to improve model accuracy for a given layer, Q-Rater compensates for bias terms of output activations.
Such selective application of bias correction implies that reducing quantization error on layer outputs (described in Figure~\ref{fig:quant_strategy}) may not reduce model accuracy drop.

Figure~\ref{fig:comparison_bias_corr} presents incremental quantization results using ResNet-32 on CIFAR-10 with different bias correction schemes (when weights and activations are quantized by using RTN and MSE clipping).
When combined with RTN and MSE clipping, bias correction is effective for initial layers in Figure~\ref{fig:comparison_bias_corr}.
Then, model accuracy is dropped sharply for a few final layers.
On the other hand, interestingly, selective bias correction by Q-Rater offers significantly improved test accuracy throughout the entire layers.
Hence, searching for quantization schemes differently for each layer based on training accuracy monitoring is effective for bias correction as well.
Interestingly, as shown in Appendix C, more layers require bias correction as $q$ decreases while the locations of such layers seem to be random.
It is also interesting that MobileNetV2 demands bias corrections for layers close to inputs and outputs.
Refer to Appendix C for experimental results using ResNet-18 and MobileNetV2 on ImageNet.

\subsection{Combination effects on Q-Rater operations}
So far, we studied Q-Rater operations (i.e., rounding, clipping, and bias correction) individually. 
Now, we combine those Q-Rater operations to see combination effects.
For experiments, as a reference quantization scheme, weights and activations are quantized by using RTN and MSE-based clipping without bias correction.
Then each quantization operation is replaced with that of Q-Rater.
We perform a grid search to find $\gamma_n \in [-1,1]$, $\gamma_s \in [0,1]$, and $\gamma_c \in [0,1]$ (by monitoring training task loss) when the search resolutions are 0.1, 0.25, and 0.1, respectively.
The entire Q-Rater procedures are described in Algorithm 1 (Appendix E).
When Q-Rater operations are combined, clipping is performed first and rounding is followed.
After weights and activations are quantized by $\gamma_n, \gamma_s$, and $\gamma_c$, selective bias correction for Q-Rater is conducted.
Note that we restrict our interests to layer-wise and symmetric quantization that is efficient for inference but challenging in terms of maintaining test accuracy.
In the case of the ImageNet dataset, we choose 20K samples only for fast evaluations of the model.

Table~\ref{table:ablation_study_final_results} presents top-1 test accuracy of ResNet-32 (on CIFAR-10), ResNet-18 (on ImageNet), and MobileNetV2 (on ImageNet) when quantized by selected previous methods and by Q-Rater.
Previous methods include three different clipping methods (i.e., MSE, ACIQ, and KL) and RTN without bias correction (as introduced in \cite{OCS}).
Note that all three individual Q-Rater operations outperform previous methods.
We also observe that combining multiple Q-Rater operations can substantially improve test accuracy for low-bit quantization.

\begin{table*}[h]
\small
\caption{Top-1 test accuracy (\%) comparison results. Q-Rater performs clipping (using $\gamma_c$), rounding (2nd-order using $\gamma_n$ and $\gamma_s$), and selective bias correction sequentially for a layer.}
\label{table:final_experimental_results}
    \vskip 0.15in
    \centering
    \begin{threeparttable}
    \begin{tabular}{ccc|ccc|cc}
    \hline
    \mr{2}{Dataset}                   & \mr{2}{Model\\(Full Acc.)}                                 & \mr{2}{W/A\\bits} & \mr{2}{MSE} & \mr{2}{OCS\tnote{1}\\(+MSE)}   & \mr{2}{Bit-Split\tnote{2}}      & \multicolumn{2}{c}{Q-Rater (\ding{202}+\ding{203}+\ding{204})} \\ \cline{7-8}
                              &                                       &     &                     &       &                & Grid Search     & Bayesian Opt.     \\ \hline
    \mr{9}{ImageNet} & \mr{3}{ResNet-18\\(69.69)}                     & 8/8 & 69.31               & 69.38 & 69.48          & \textbf{69.43} & \textbf{69.67} \\
                              &                                       & 6/6 & 66.75               & 67.51 & 68.59          & \textbf{68.46} & \textbf{68.60} \\
                              &                                       & 4/4 & 32.16               & 34.85 & 55.18          & \textbf{59.17} & \textbf{60.77} \\ \cline{2-8} 
                              & \mr{3}{ResNet-101\\(77.79)}           & 8/8 & 77.02               & 77.18 & 77.20 & \textbf{77.10} & \textbf{77.16} \\
                              &                                       & 6/6 & 74.20               & 75.36 & 0.08\tnote{3} & \textbf{75.97} & \textbf{76.21} \\
                              &                                       & 4/4 & 19.48               & 29.76 & 0.06\tnote{3} & \textbf{64.49} & \textbf{68.24} \\ \cline{2-8} 
                              & \mr{3}{MobileNetV2\\(71.78)}         & 8/8 & 69.81               & N/A   & 70.60\tnote{4}      & \textbf{71.25} & \textbf{71.10} \\
                              &                                       & 6/6 & 36.82               & N/A   & 53.88\tnote{4}      & \textbf{67.64} & \textbf{68.20} \\
                              &                                       & 4/4 & 0.12                & N/A   & 0.28\tnote{4}      & \textbf{25.35} & \textbf{22.95} \\ \hline
    \mr{3}{CIFAR-10}  & \mr{3}{ResNet-32\\(92.63)}                    & 5/5 & 90.64               & 91.14 & 91.68          & \textbf{91.74} & \textbf{91.84} \\
                              &                                       & 4/4 & 84.46               & 86.87 & 89.50          & \textbf{90.24} & \textbf{90.15} \\
                              &                                       & 3/3 & 16.86               & 35.97 & 76.07          & \textbf{79.69} & \textbf{79.71} \\ \hline
    \end{tabular}
            \begin{tablenotes}
            \footnotesize
            \item[1] Reference code: https://github.com/cornell-zhang/dnn-quant-ocs
            \item[2] Reference code: https://github.com/wps712/BitSplit
            \item[3] Numerical instability is observed probably because of matrices of too large dimensions.
            \item[4] While depthwise convolution is not discussed in the reference, for our experiments, we revise the reference code such that depthwise convolution layers are quantized in a way to quantize fully-connected layers.
        \end{tablenotes}
    \end{threeparttable}
    \vskip -0.1in
\end{table*}

\section{Experimental results}
The success of Q-Rater largely depends on the search quality of hyper-parameters $\gamma_n$, $\gamma_s$, and $\gamma_c$.
Even though a coarse-grained grid search method can produce impressive results as shown in Table~\ref{table:ablation_study_final_results}, as a way of fine-tuning hyper-parameters, we adopt Bayesian Optimization\footnote{We use a publicly available code in \cite{Nogueira2014Bayesian}.} (BO), one of the automated techniques to search parameter spaces. 
After a brief introduction of the BO technique, we introduce the outlier channel split (OCS) \cite{OCS} and bit-split \cite{bit-split} as examples minimizing quantization error on weights and layer outputs, respectively.
Then, we show comparison results on model accuracy.

\paragraph{Bayesian optimization}
 For a given dataset $D$ and a given parameter vector $\vh{=}\{\gamma_c\}$ or $\vh{=}\{\gamma_n, \gamma_s\}$, BO aims to find an optimal $\vh^{*}$ which maximizes an evaluation function $f(\vh, D)$ for a network. Four times processes for each layer are required in our suggested algorithm (refer to Algorithm 1 in Appendix). To avoid over-fitting of quantization parameters to a test dataset, a training dataset or a subset of the training dataset is used for evaluation at each step. Using grid search results as initial search space, in our experiments, BO explores additional 50 hyper-parameter sets.
 
 \paragraph{Outlier channel split}
 OCS represents recent efforts to minimize quantization error on weights.
 Halved outliers in weights due to OCS without model structure modifications enable aggressive clipping that plays a pivotal role in uniform quantization.
 It is reported that OCS improves MSE, ACIQ, and KL-based clipping noticeably especially for low-bit quantization of weights \cite{OCS}.

 \paragraph{Bit-split}
 We introduce the bit-split technique as one of the attempts to minimize reconstruction error (i.e., quantization error on layer outputs) for post-training uniform quantization.
 Bit-split formulates quantization as an optimization problem where quantization-related parameters are computed in an iterative fashion to solve a complicated objective function.
 While partial analytical solutions are obtained during iterative procedures, matrix multiplications can entail large $\mX$ matrices whose dimensions increase with the number of input samples and the size of input features.

 \paragraph{Comparison results}
 Table~\ref{table:final_experimental_results} presents comparison results on top-1 test accuracy of ResNet-18, ResNet-101 (representing large depth models), and MobileNetV2 (challenging to be quantized mainly due to depth-wise layers) on ImageNet dataset, and ResNet-32 on CIFAR-10 dataset.
 \textbf{For ImageNet, Q-Rater does not need the entire dataset.}
 Instead, we use only 20K random samples to estimate the loss function (see Appendix D for the relationship between the number of samples and quantization quality).
 For Q-Rater with grid search, we use the same search resolutions of the previous section (i.e., 0.1 for $\gamma_n$, 0.25 for $\gamma_s$, and 0.1 for $\gamma_c$).
 We observe that even for layer-wise and symmetric quantization we choose for this work, OCS indeed enhances MSE.
 Then, bit-split outperforms OCS except for the ResNet-101 model where we observe numerical instability during computations (probably because of too complicated analytical equations that can be potentially solved by higher precision such as double precision).
 Note that since we assume layer-wise quantization, bit-split entails a lot more complicated computations compared to channel-wise quantization that is chosen in \cite{bit-split}.
 For all models in the table, Q-Rater presents the best test accuracy while BO slightly exceeds a grid search for most cases.


\section{Conclusion}

In this paper, we propose a new post-training uniform quantization technique, called Q-Rater, which does not depend on convex optimizations.
While previous works mainly compute hyper-parameters for quantization to minimize quantization errors on weights or layer outputs, we suggest that we can perform a grid search of quantization hyper-parameters by evaluating corresponding training loss.
For each layer, hyper-parameters for clipping and rounding are searched, and then bias correction is conducted selectively.
Through layer-wise incremental optimization for hyper-parameters, Q-Rater presents significantly higher model accuracy for various CNN models even assuming a simple layer-wise and symmetric quantization format.

\bibliography{QRator}
\bibliographystyle{icml2021}

\appendix
\onecolumn
\setcounter{figure}{8}
\setcounter{table}{2}



\section{Evaluation of Q-Rater Rounding Scheme}

In this section, we provide comprehensive experimental results to demonstrate the effectiveness of the proposed Q-rater rounding scheme. We show how the 1st-order and the 2nd-order Q-rater rounding methods divide the entire range by using different $\gamma_n$. And then, we evaluate the proposed scheme on various models. For the evaluation, pre-trained ResNet32 on CIFAR-10 dataset, ResNet18 and MobileNetV2 on ImageNet dataset are used. We present the experimental results of the 1st-order and the 2nd-order rounding schemes. 

\subsection{Q-Rater Range Division Examples} 

\begin{figure}[h]
    \centering
    \includegraphics[width=0.43\linewidth]{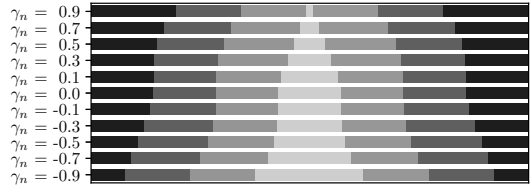}
    \caption{Examples of unequal range division by the 1st-order rounding scheme with various $\gamma_n$. 
    }
    \label{fig:1st-range}
\end{figure}

\begin{figure}[h]
\centering
    \begin{subfigure}[t]{0.43\textwidth}
    \centering
    \includegraphics[width=\linewidth]{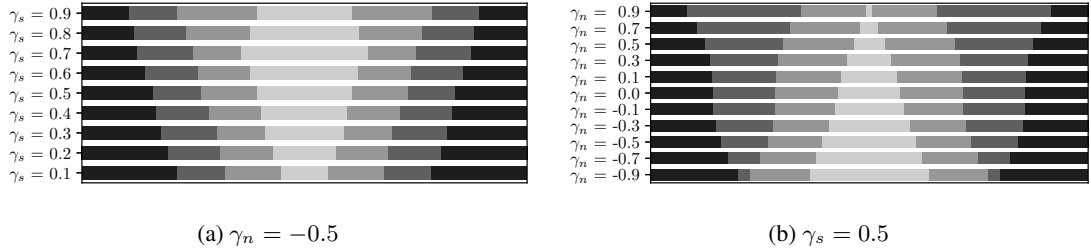}
    \caption{$\gamma_n = -0.5$}
    \end{subfigure}
    \begin{subfigure}[t]{0.43\textwidth}
    \centering
    \includegraphics[width=\linewidth]{eps/nonlinear_2nd_exp_s0.50.eps}
    \caption{$\gamma_s = 0.5$}
    \end{subfigure}
    \caption{Examples of unequal range division by the 2nd-order rounding scheme with various $\gamma_n$ and $\gamma_s$. 
    }
    \label{fig:2nd-range}
\end{figure}

Figure~\ref{fig:1st-range} and Figure~\ref{fig:2nd-range} show how the range is divided by the 1st-order and the 2nd-order Q-Rater rounding. If $\gamma_n$ is zero, then Q-Rater is exactly the same as the conventional Rounding-to-Nearest (RTN) scheme in both cases. With the 1st-order Q-Rater rounding, the size of range monotonically increases or decreases from the zero. For example, the range always decreases as it goes further away from the zero when $\gamma_n > 0$. Meanwhile, the range always increases as it goes further away from the zero when $\gamma_n < 0$. With the 2nd-order Q-Rater rounding, $\gamma_s$ decides where to start increase or decrease. 



\clearpage

\subsection{ResNet-32 on CIFAR-10}

\begin{figure}[h]
    \centering
    \includegraphics[width=0.8\linewidth]{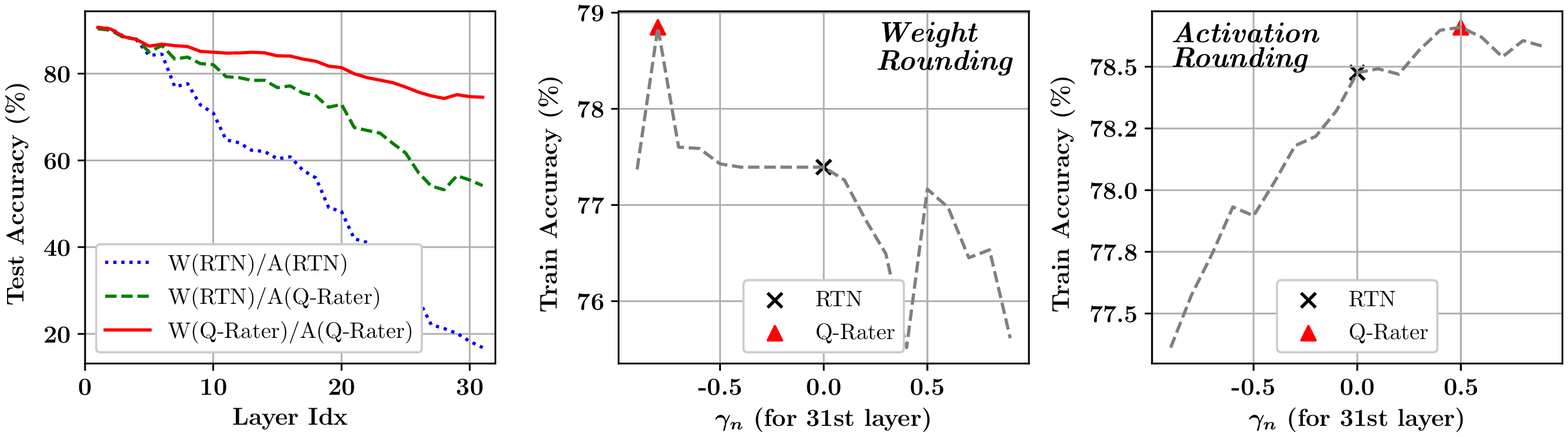}
    \caption{Comparison between Rounding-to-Nearest (RTN) and the 1st-order Q-Rater rounding. 3-bit quantization is performed from the second layer to the last layer. MSE minimization is used for the clipping method.}
    \vspace{-0.1in}
    \label{fig:1st-res32}
\end{figure}

Figure~\ref{fig:1st-res32} compares the Rounding-to-Nearest (RTN) scheme and the 1st-order Q-Rater rounding scheme using ResNet32. 
As quantization is performed incrementally per layer, test accuracy continues to drop for both quantization schemes. Note that Q-Rater rounding yields less accuracy drop compared to the RTN throughout layers. It is more effective when Q-Rater is applied to both the weights and activation, though the case when Q-Rater is applied to the activations only also outperforms the RTN. The second and the third plots show train accuracy by different $\gamma_n$ values used in the weight rounding and the activation rounding, respectively. If $\gamma_n$ is 0.0, the result is the same as the RTN. Those two plots with various $\gamma_n$ indicate that the RTN does not guarantee the best accuracy and there exist opportunities to find other parameters that provide better accuracy. The 1st-order Q-Rater rounding scheme searches a particular $\gamma_n$ that exhibits the best training accuracy. 

%

\clearpage

\subsection{ResNet-18 on ImageNet}

\begin{figure}[h]
\centering
    \begin{subfigure}[t]{0.8\textwidth}
    \centering
    \includegraphics[width=\linewidth]{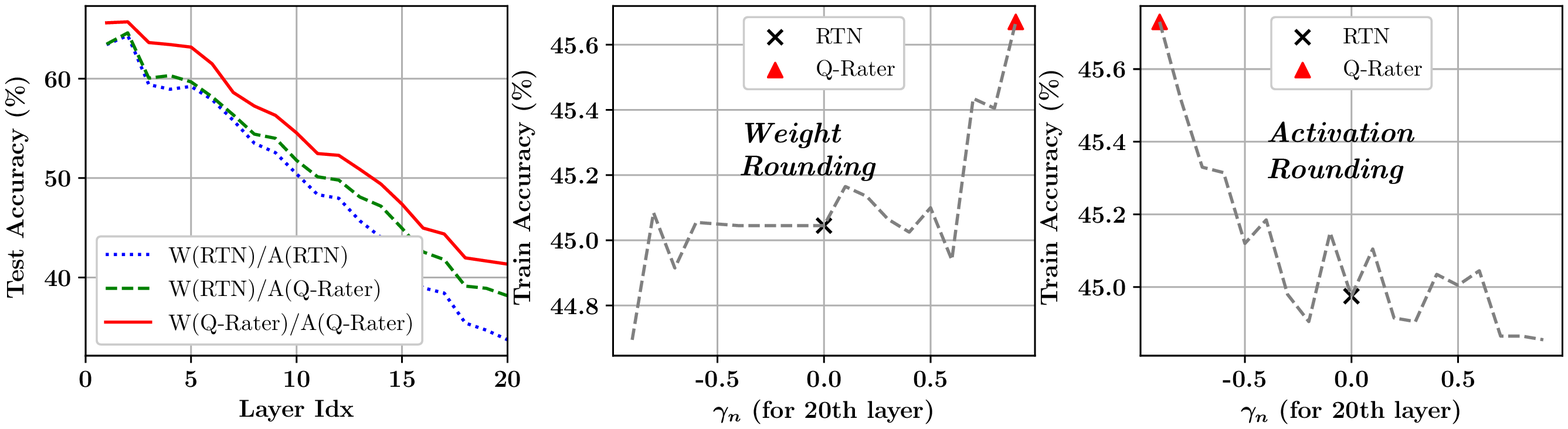}
    \caption{1st-order rounding scheme}
    \end{subfigure}
    \begin{subfigure}[t]{0.8\textwidth}
    \centering
    \includegraphics[width=\linewidth]{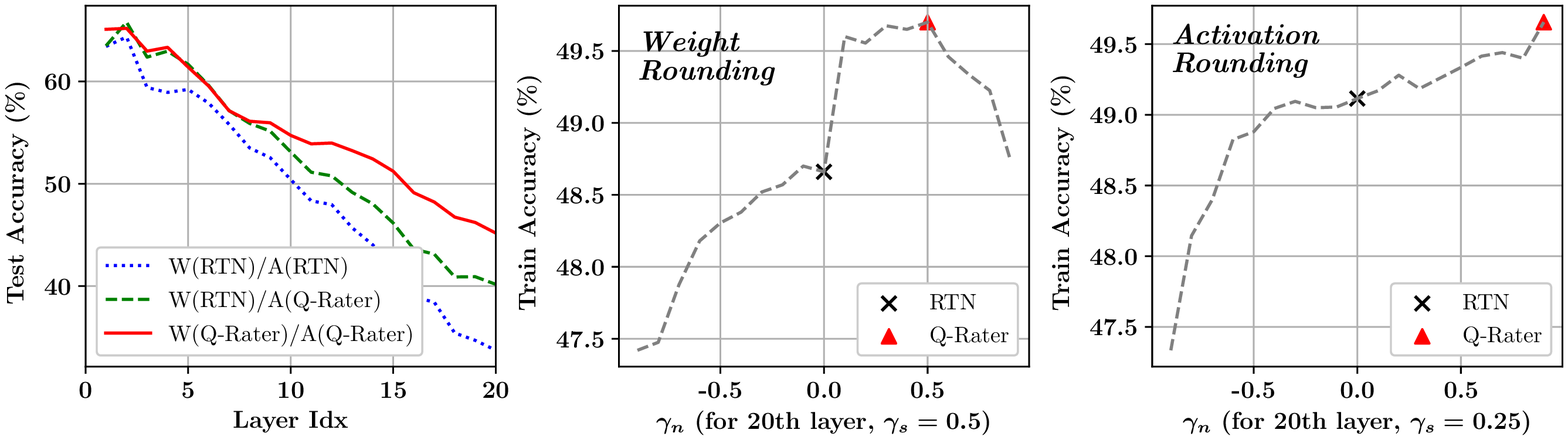} 
    \caption{2nd-order rounding scheme}
    \end{subfigure}
    \vspace{-0.1in}
    \caption{Comparison of Rounding-to-Nearest(RTN) and the 1st-order and the 2nd-order Q-Rater. 4-bit quantization is performed to all layers including the first and the last layer. MSE minimization is used for the clipping method. }
    \label{fig:resnet18}
\end{figure}


Figure~\ref{fig:resnet18} compares the Rounding-to-Nearest (RTN) scheme and the 1st-order and the 2nd-order Q-Rater rounding scheme in ResNet-18. The ResNet-18 model used in the quantization is trained by using the ImageNet dataset. As illustrated in Figure~\ref{fig:1st-res32}, Q-Rater is more tolerant to the accuracy drop by quantization compared to the RTN. Specifically, Q-Rater shows the best performance when the 2nd-order rounding scheme is applied to both the weights and the activations. 
The second and the third plots in Figure~\ref{fig:resnet18}(a) and (b) show train accuracy by different $\gamma_n$. For the 2nd-order rounding scheme, $\gamma_s$ is fixed to the value depicted in the figure. As shown in the figure, train accuracy varies by $\gamma_n$, and RTN does not necessarily result in the best train accuracy for both weight rounding and activation rounding. By estimating train accuracy, Q-Rater finds $\gamma_n$ and $\gamma_s$ to achieve the best accuracy. 

\vspace{-0.1in}
\begin{figure}[h]
    \centering
    \includegraphics[width=0.75\linewidth]{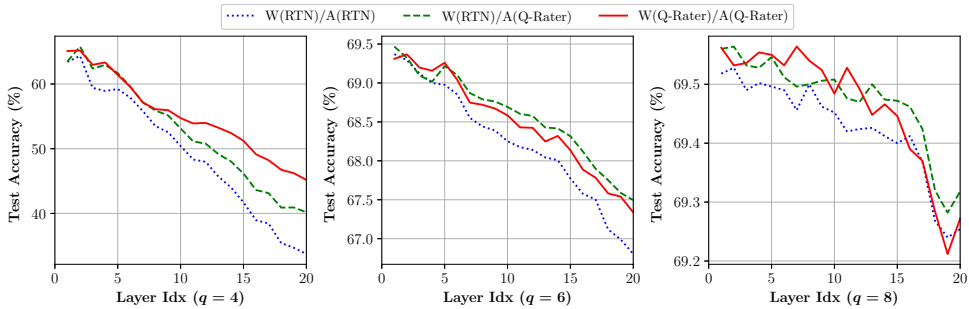}
    \caption{Comparison of Rounding-to-Nearest (RTN) and the 2nd-order Q-Rater by using the different number of quantization bits. Quantization is performed from the second to the last layer. MSE minimization is used for the clipping.}
    \label{fig:resnet18-by-bits}
\end{figure}
\vspace{-0.1in}

Figure~\ref{fig:resnet18-by-bits} compares the Rounding-to-Nearest (RTN) scheme and the 2nd-order Q-Rater rounding scheme using the ResNet18 model. From left to right, $q=4$, $q=6$, or $q=8$ is used to quantize the weights and the activations. With 8-bit quantization, both RTN and Q-Rater show nearly full accuracy. In the case of 6-bit quantization, the difference between W(RTN)/A(Q-Rater) and W(Q-Rater)/A(Q-Rater) is less than 0.1\%. When $q=4$ is used as a more extreme environment, W(Q-Rater)/A(Q-Rater) significantly outperforms the conventional RTN scheme. 

\clearpage

\subsection{MobileNetV2 on ImageNet}
\begin{figure}[h]
    \centering
    \begin{subfigure}[t]{0.8\textwidth}
    \centering
    \includegraphics[width=\linewidth]{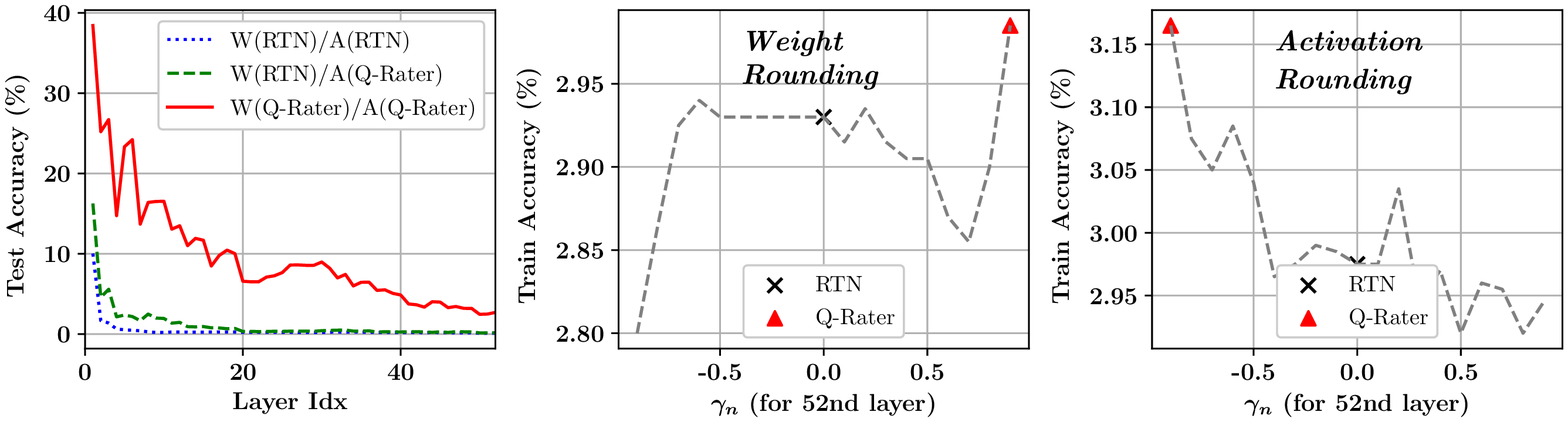}
    \caption{1st-order rounding scheme}
    \end{subfigure}
    \begin{subfigure}[t]{0.8\textwidth}
    \centering
    \includegraphics[width=\linewidth]{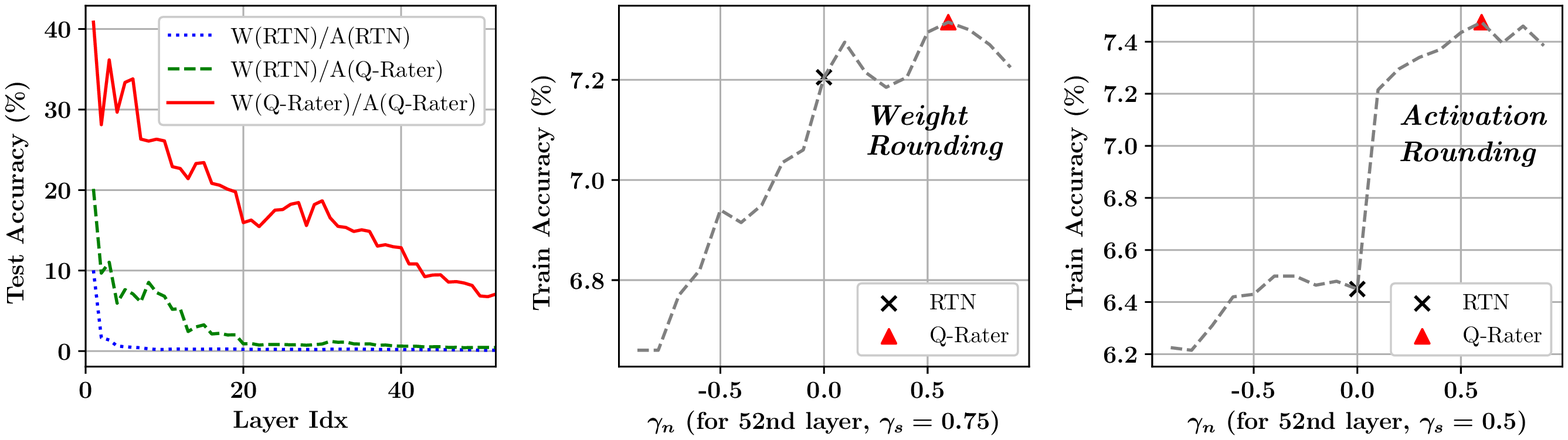} 
    \caption{2nd-order rounding scheme}
    \end{subfigure}
    \caption{Comparison of Rounding-to-Nearest (RTN) and the 1st-order and the 2nd-order Q-Rater. 4-bit quantization is performed from the second to the last layer. MSE minimization is used for the clipping method.}
    \label{fig:mobilenet}
\end{figure}

Figure~\ref{fig:mobilenet} compares the Rounding-to-Nearest (RTN) scheme and the 1st-order and the 2nd-order Q-Rater rounding scheme using MobileNetV2, trained on the ImageNet dataset. Similar to our observations on other models, Q-Rater is more tolerant in the accuracy drop by quantization compared to the RTN, and it shows the best performance when the 2nd-order rounding scheme is applied to both the weights and the activations. 
The second and the third plots in Figure~\ref{fig:mobilenet}(a) and (b) show train accuracy by different $\gamma_n$. For the 2nd-order rounding scheme, $\gamma_s$ is fixed to the value depicted in the figure. As shown in the figure, train accuracy varies by $\gamma_n$ and RTN does not necessarily result in the best train accuracy in both cases. 
Depending on the measured train accuracy, Q-Rater finds $\gamma_n$ and $\gamma_s$ to achieve the best accuracy. 


\begin{figure}[h]
    \centering
    \includegraphics[width=0.75\linewidth]{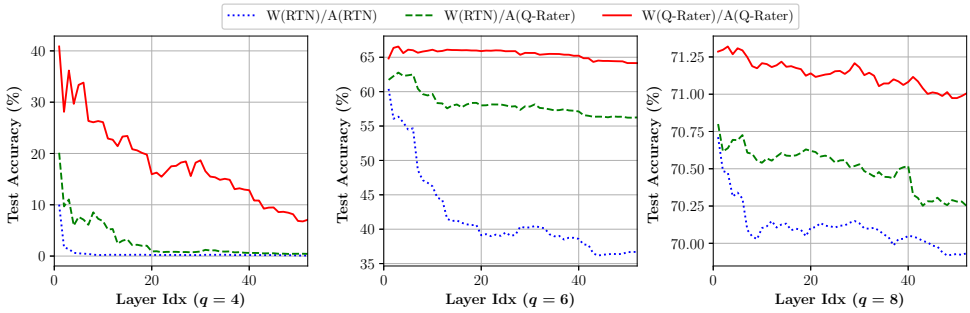}
    \caption{Comparison of Rounding-to-Nearest (RTN) and the 2nd-order Q-Rater by using the different number of quantization bits. Quantization is performed from the second to the last layer. MSE minimization is used for the clipping. }
    \label{fig:mobilenet-by-bits}
\end{figure}

Figure~\ref{fig:mobilenet-by-bits} compares the Rounding-to-Nearest (RTN) scheme and the 2nd-order Q-Rater rounding scheme in the MobileNetV2 model. From left to right, $q=4$, $q=6$, or $q=8$ is used to quantize the weights and the activations. For 8-bit quantization, W(Q-Rater)/A(Q-Rater) shows improvement over the others and produces nearly full accuracy while other schemes do not. In the case of 6-bit quantization, Q-Rater significantly improves the accuracy even in the case when Q-Rater is applied only to the activations. When $q=4$ is used as a more extreme environment, W(Q-Rater)/A(Q-Rater) outperforms the conventional RTN scheme as well. 

\subsection{Comparison of the 1st order and 2nd order schemes.}

\begin{table*}[h]
    \centering
    \begin{threeparttable}
        \caption{Comparison of top-1 accuracy (\%) with the proposed Q-Rater. 
        }
        \label{tab:rounding-summary}
        \begin{tabular}{ccc|ccc|ccc}
        \hline
        \mr{2}{Dataset}  & \mr{2}{Model}                         & \mr{2}{W/A}          & \multicolumn{3}{c|}{1st-order}  & \multicolumn{3}{c}{2nd-order} \\ \cline{4-9}
                         &                                       &                      & \ding{202}\tnote{1} & \ding{202}+\ding{203}\tnote{3}  & \ding{202}+\ding{203}+\ding{204}\tnote{4} & \ding{202}\tnote{2} & \ding{202}+\ding{203}\tnote{3}     & \ding{202}+\ding{203}+\ding{204}\tnote{4}  \\ \hline
        \mr{6}{ImageNet} & \mr{3}{ResNet-18\\(69.69)}            & 8/8                  & 69.53         & 69.51   & 69.51                 & 69.27         & 69.33      & 69.43 \\
                         &                                       & 6/6                  & 67.19         & 67.90   & 68.44                 & 67.19         & 67.95      & 68.46 \\
                         &                                       & 4/4                  & 41.34         & 54.71   & 61.06                 & 41.34         & 54.87      & 59.17 \\ \cline{2-9}
                         & \mr{3}{MobileNetV2\\(71.78)}          & 8/8                  & 69.86         & 70.84   & 71.23               & 71.01         & 70.92      & 71.25 \\
                         &                                       & 6/6                  & 55.17         & 63.97   & 67.16               & 64.15         & 65.33      & 67.64 \\
                         &                                       & 4/4                  &  2.70         & 13.46   & 14.61               &  7.08         & 15.58      & 25.35 \\ \hline
        \mr{3}{CIFAR-10}  & \mr{3}{ResNet-32\\(92.63)}           & 5/5                  & 91.54         & 91.73   & 91.77               & 91.70         & 91.93      & 91.74 \\
                         &                                       & 4/4                  & 89.02         & 89.02   & 89.69                 & 89.03         & 89.71      & 90.24 \\
                         &                                       & 3/3                  & 74.56         & 74.56   & 76.76                 & 75.74         & 77.35      & 79.69 \\ \hline
        \end{tabular}
        \begin{tablenotes}
            \footnotesize
            \item[1] 1st-order rounding scheme of Q-Rater
            \item[2] 2nd-order rounding scheme of Q-Rater
            \item[3] Clipping method of Q-Rater
            \item[4] Bias Correction of Q-Rater
        \end{tablenotes}
    \end{threeparttable}
\end{table*}

Table~\ref{tab:rounding-summary} summarizes the results of the 1st-order and the 2nd-order rounding scheme. The results with the 2nd-order scheme are the same as in Table 1. In general, the 2nd-order scheme results in better accuracy than the 1st-order scheme. Combined with the clipping method and the bias correction of Q-Rater, the accuracy can be further improved.  

\clearpage
\section{Evaluation of Q-Rater Clipping Methods}
In this section, we evaluate the effectiveness of the proposed Q-Rater clipping methods over the ResNet-18 and the MobileNetV2 models. Both models used in the evaluation are trained on ImageNet dataset. In the paper, we introduced $\gamma_c$ as a new hyper-parameter for the clipping. Note that $\gamma_c$ ranges between 0.0 and 1.0. For the conventional scheme, we choose MSE minimization, ACIQ, and KL divergence minimization. 
For more details on the conventional methods, please refer to Section 5.2. 

\subsection{ResNet-18}
\begin{figure}[h]
    \centering
    \includegraphics[width=0.85\linewidth]{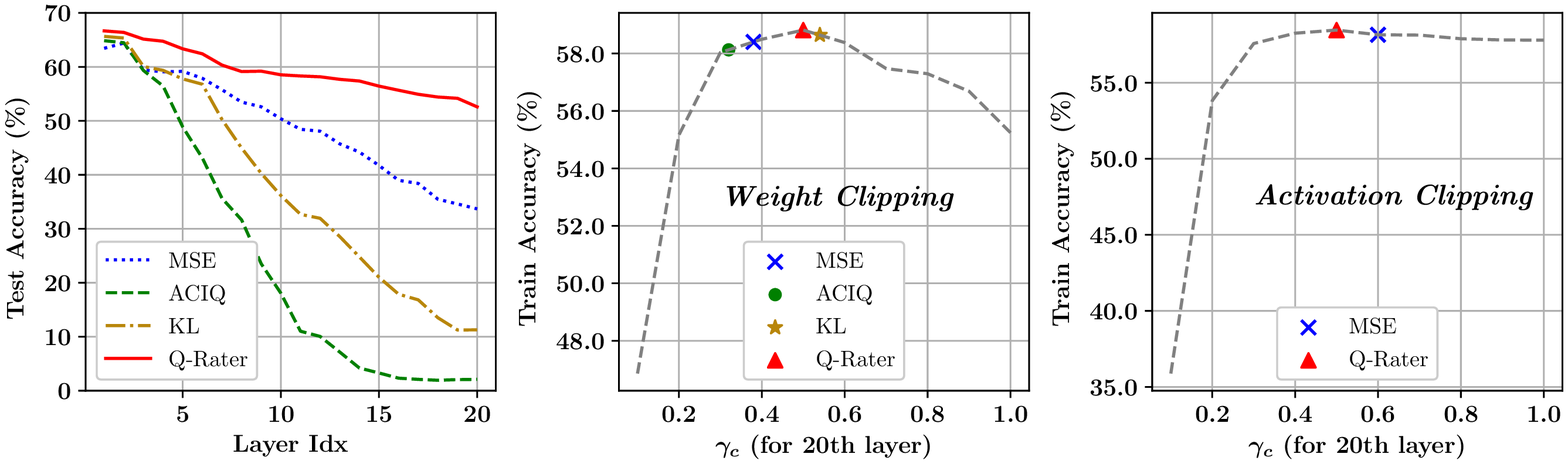}
    \caption{Comparison of different clipping methods with RTN rounding scheme. For quantization, 4 bits are used for both weights and activations. }
    \label{fig:resnet18-clip}
\end{figure}
\vspace{-0.1in}

Figure~\ref{fig:resnet18-clip} compares different clipping methods with RTN rounding scheme in ResNet-18. The clipping and the quantization are performed per-layer incrementally as in the previous experiments. Thus, test accuracy drops as the layer index increases as in the leftmost figure. However, the proposed Q-Rater clipping method is remarkably tolerant in the accuracy drop compared to other schemes. Test accuracy with Q-Rater clipping and RTN rounding results in $55.48\%$. 
The second and the third plots show training accuracy by different $\gamma_c$ applied in the weight clipping and the activation clipping respectively. Each clipping method calculates different $\gamma_c$ with their own algorithms, but that does not necessarily result in the best accuracy. Based on the training accuracy, Q-Rater finds $\gamma_c$ to achieve the best accuracy. 

\clearpage

\subsection{MobileNetV2}
\begin{figure}[h]
    \centering
    \includegraphics[width=0.85\linewidth]{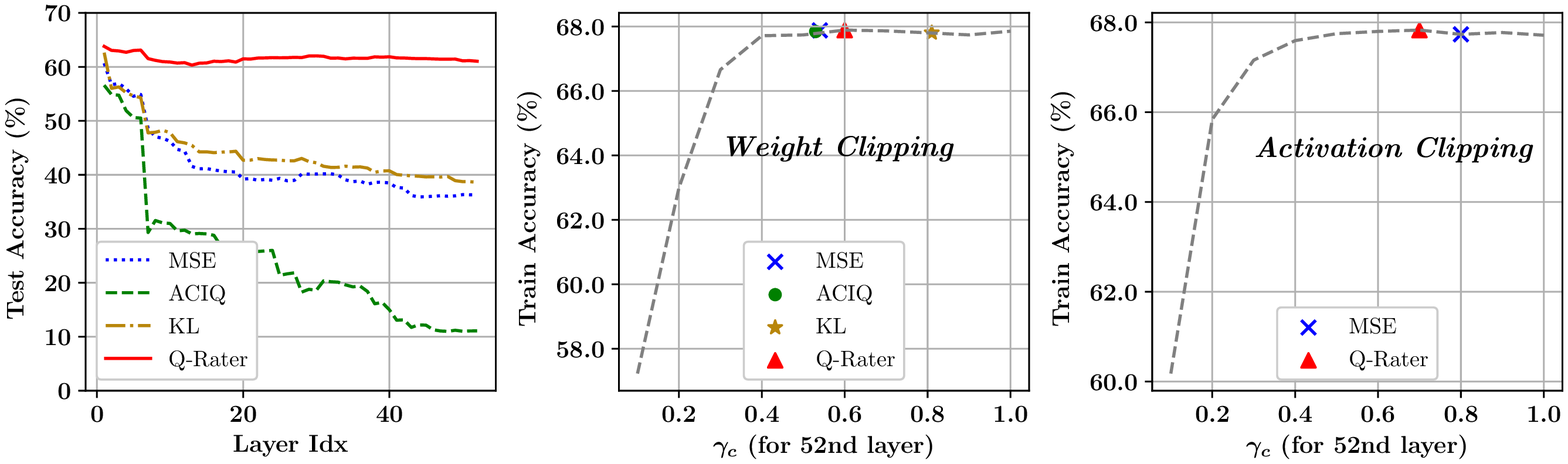}
    \caption{Test accuracy comparison using MSE, ACIQ, KL, or our proposed clipping method with RTN rounding scheme for weights and activations of MobileNetV2 model on ImageNet ($q=6$ for both weights and activations).}
    \label{fig:mobilenet-clip}
\end{figure}
\vspace{-0.1in}

Figure~\ref{fig:mobilenet-clip} compares different clipping methods with RTN rounding scheme in MobileNetV2. The clipping and the quantization are performed per-layer incrementally as in the previous experiments. The first plot indicates that Q-Rater significantly outperforms other clipping methods. The second and the third plots show train accuracy by different $\gamma_c$ applied in the weight clipping and the activation clipping respectively. Note that Q-Rater clipping is applied until the 52nd layer. Only the last layer of the model is clipped with different clipping methods in the second and the third plot. Since the accuracy drop by different clipping schemes mostly occurs at the early layers, the difference of train accuracy among those clipping methods seems comparatively small.

\clearpage

\section{Evaluation of Q-Rater Bias Correction Methods}

In this section, we evaluate the effectiveness of the proposed Q-Rater bias correction methods over the ResNet-18 and the MobileNetV2 models. Both models used in the evaluation are trained on the ImageNet dataset. In Q-Rater, bias correction is applied per layer only when it improves train accuracy. 

\subsection{ResNet-18}
\begin{figure}[h]
    \centering
    \includegraphics[width=0.5\linewidth]{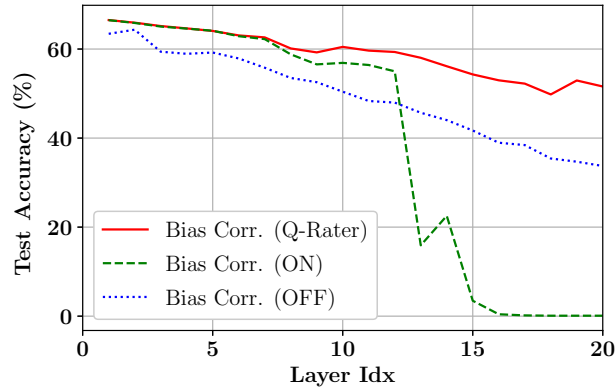}
    \caption{Test accuracy of ResNet-18 model on ImageNet when bias correction is always applied, never applied, or selectively applied by Q-Rater. Weights and activations are quantized using $q=4$ incrementally with RTN and MSE clipping.}
    \label{fig:resnet18-bc}
\end{figure}

Figure~\ref{fig:resnet18-bc} shows the test accuracy of ResNet-18 model trained on the ImageNet dataset. Interestingly, ResNet-18 shows the worst performance when bias correction is applied to all layers. Bias correction is supposed to compensate for the errors from quantization. However, the result indicates that minimizing quantization errors by bias correction may not lead to better performance. The result indicates Q-Rater effectively achieves the best performance by selectively enabling the bias correction per layer. 

\subsection{MobileNetV2}
\begin{figure}[h]
    \centering
    \includegraphics[width=0.5\linewidth]{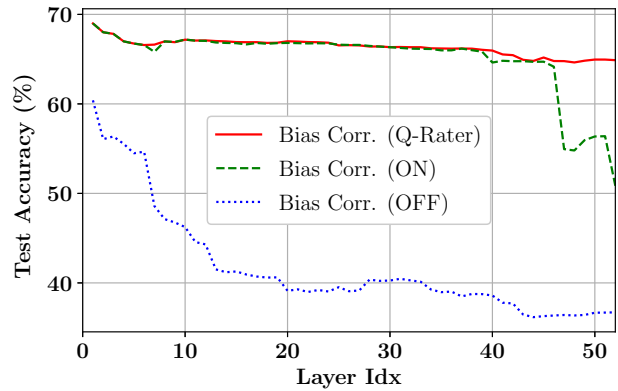}
    \caption{Test accuracy of MobileNetV2 model on ImageNet when bias correction is always applied, never applied, or selectively applied by Q-Rater. Weights and activations are quantized using $q=6$ incrementally with RTN and MSE clipping.}
    \label{fig:mobilenet-bc}
\end{figure}

Figure~\ref{fig:mobilenet-bc} shows the test accuracy of MobileNetV2 model trained on the ImageNet dataset. Although applying bias correction is effective on many layers, accuracy drops abruptly at the 40th layer and the 46th layer. We observed that weight values are biased and have high variance per channel at those layers. For example, the entire weights in some channels are negative, while the entire weights in other channels are all positive. In such cases, applying bias correction only adds errors to both channels.  The result indicates Q-Rater effectively achieves the best performance by selectively enabling the bias correction per layer. 

\subsection{Bias Correction Selection Results}
\begin{figure}[h]
\begin{center}
\begin{subfigure}[t]{0.4\textwidth}
    \centering
    \includegraphics[width=\textwidth]{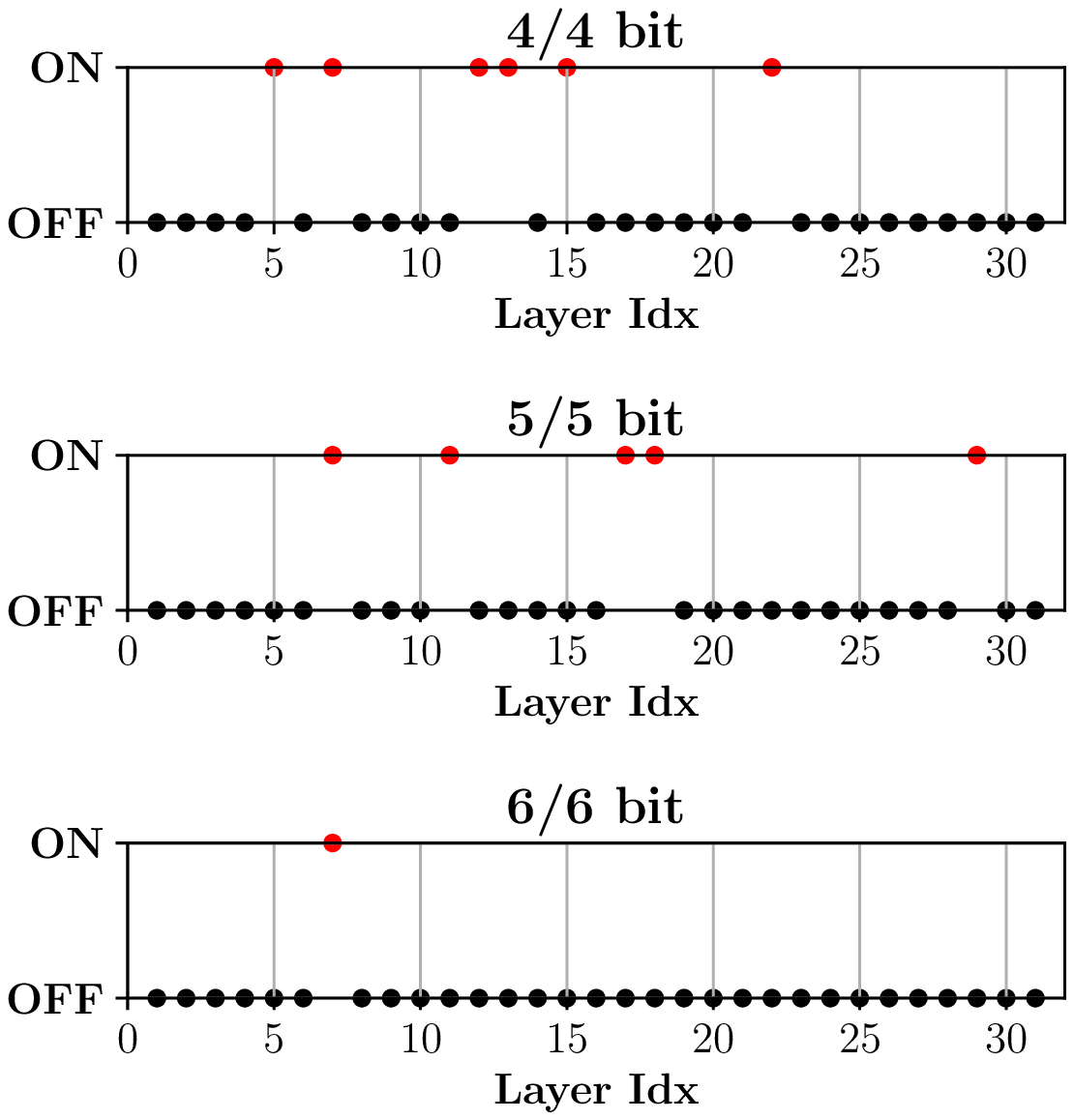}
    \caption{ResNet-32 on CIFAR-10}
    \label{fig:onoff_r32}
\end{subfigure}
\begin{subfigure}[t]{0.4\textwidth}
    \centering
    \includegraphics[width=\textwidth]{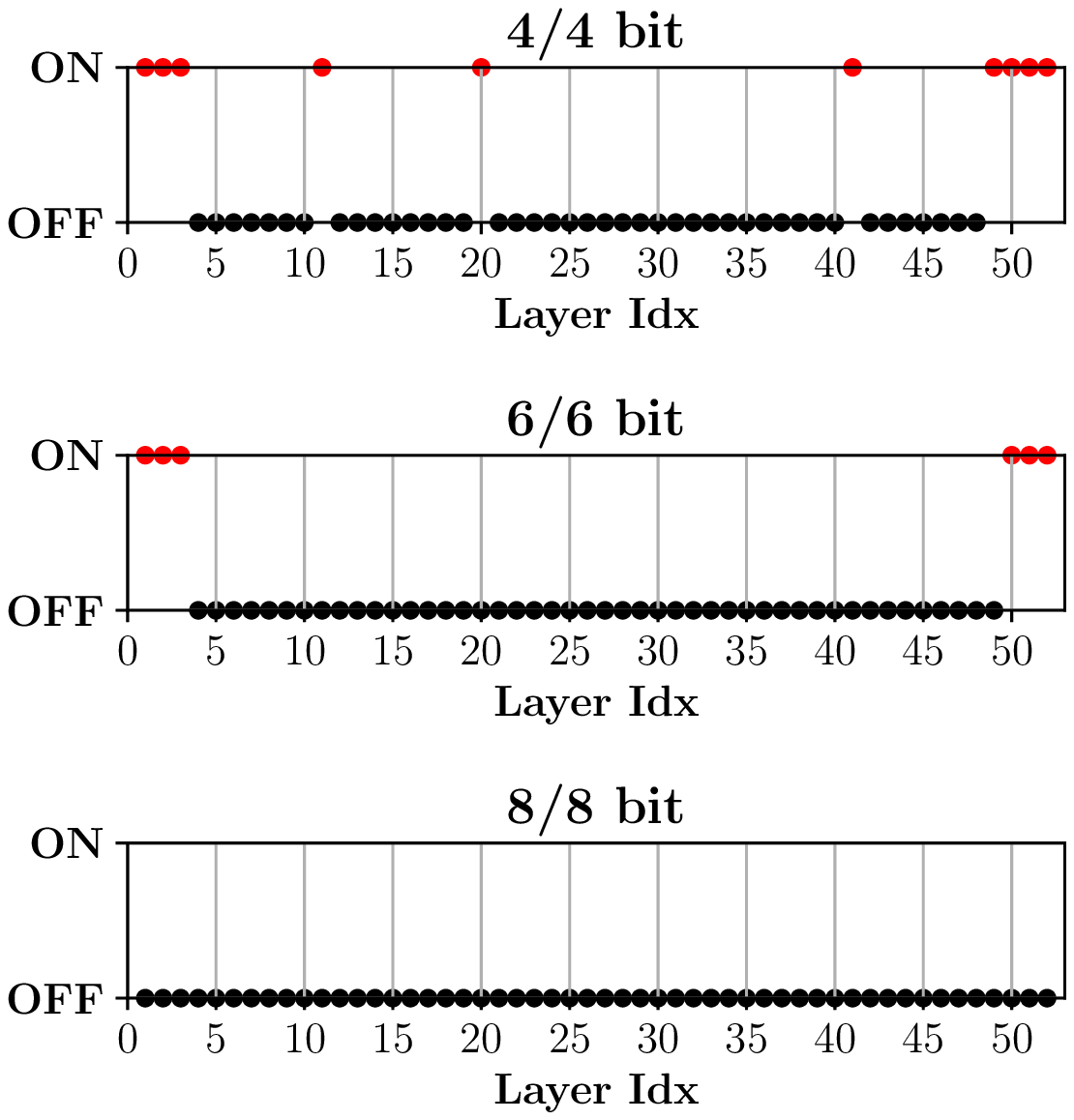}
    \caption{MobileNetV2 on ImageNet}
    \label{fig:onoff_m2}
\end{subfigure}
\end{center}
\begin{center}
\begin{subfigure}[t]{0.4\textwidth}
    \centering
    \includegraphics[width=\textwidth]{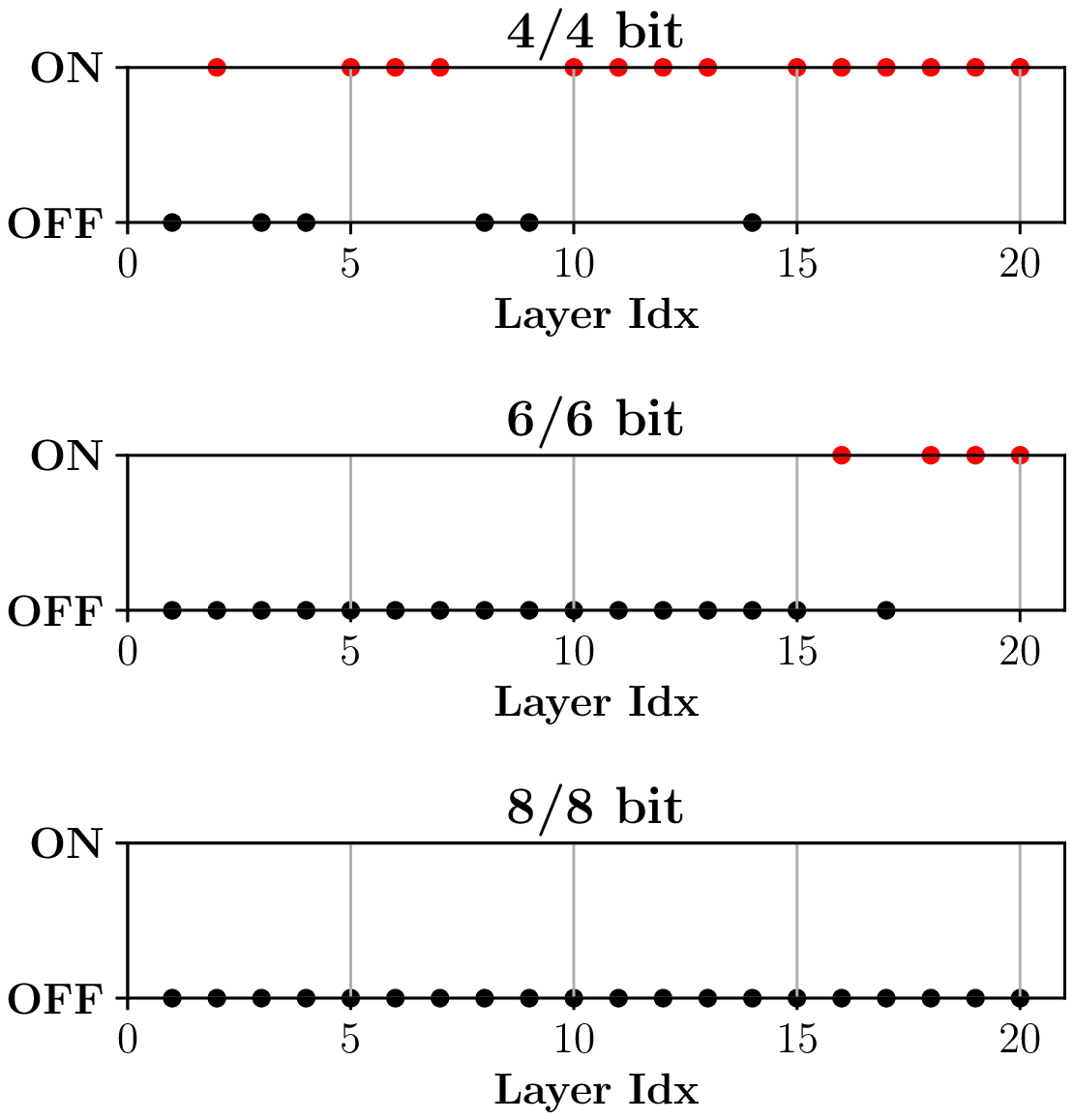}
    \caption{ResNet-18 on ImageNet}
    \label{fig:onoff_r18}
\end{subfigure}
\begin{subfigure}[t]{0.4\textwidth}
    \centering
    \includegraphics[width=\textwidth]{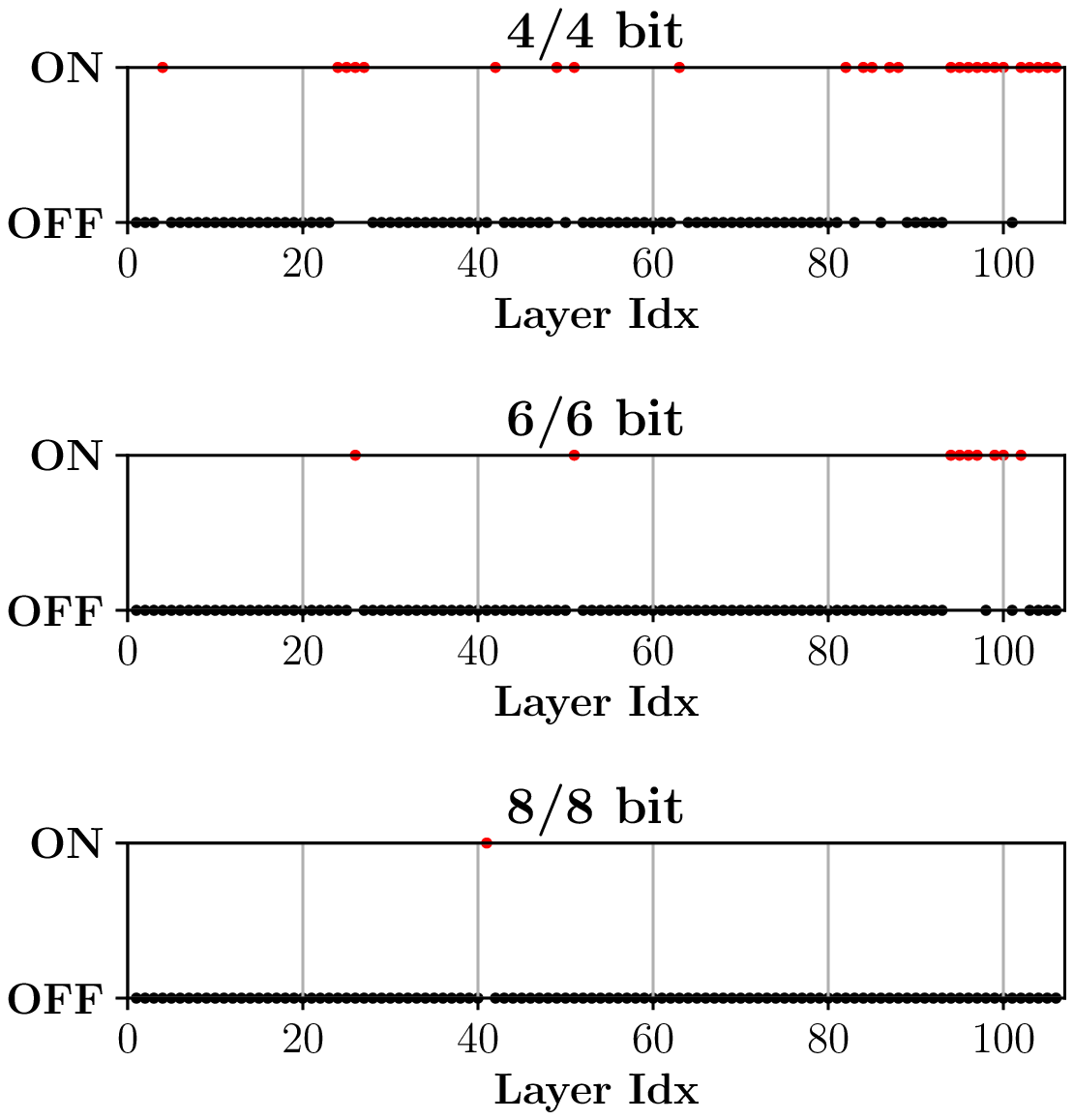}
    \caption{ResNet-101 on ImageNet}
    \label{fig:onoff_r101}
\end{subfigure}
\end{center}
\caption{Bias correction On-Off patterns on different models. }
\label{fig:onoff}
\end{figure}

Figure \ref{fig:onoff} presents which layers have been selected for the bias correction. Overall, our results do not show specific patterns on bias correction selections except for two observations: 1) when a model is quantized into lower bits, bias correction seems to be more effective because bias correction can compensate accumulated errors by quantization (e.g., bias correction is seldomly applied to 8-bit quantized models.) and 2) For quantized MobileNetV2 with 4/4 and 6/6 bits, it is effective to apply bias correction to a few layers close to inputs or outputs.

\section{The Impact of Calibration Set Size on ResNet-18 Quantization}
\begin{figure}[h]
    \centering
    \includegraphics[width=0.5\linewidth]{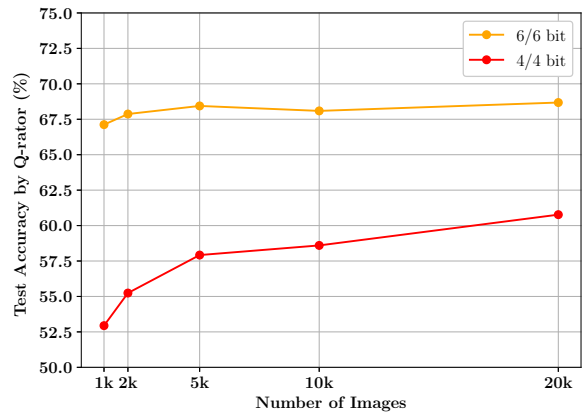}
    \caption{Test accuracy of ResNet18 model on ImageNet $N$k is used. }
    \label{fig:app_datasize}
\end{figure}

For the experimental results in the main manuscript, we used 20k ImageNet dataset which consists of 20 randomly picked images per class from the original ImageNet dataset.  
Since using the whole dataset for hyper-parameter search (e.g., for $m$, $s$) is impractical and inefficient, we can select a dataset of small size if such a small dataset can exhibit a similar distribution of the original dataset. 
Figure~\ref{fig:app_datasize} shows the test accuracy measured on ResNet18 model quantized by 4/4 and 6/6 bits while we increase the number of images per class. We observe that our selection (20 images per class in the main manuscript) is enough to obtain high test accuracy.

\clearpage
\section{Algorithm Descriptions on Q-Rater Operations}

In Algorithm~\ref{alg:main-alg}, we provide pseudo-codes of the proposed Q-Rater algorithm. We briefly explain each function. 

\vspace{0.1in}
\noindent
{\bf Run()} Q-Rater executes Run() to quantize a model. For each layer, three functions are called to quantize weights, quantize activations, and perform bias correction. 

\vspace{0.1in}
\noindent
{\bf WeightQuant$(i)$} This function quantizes the weight values. Firstly, the original weight values are stored. And then, the function finds $\gamma_c$ through Bayesian Optimization and performs clipping. When the weight clipping is finished, the function continues finding hyper-parameters for quantization: $\gamma_n$ and $\gamma_s$. With the determined hyper-parameters, the function performs quantization. 

\vspace{0.1in}
\noindent
{\bf WeightClip$(w_{org}, \gamma_c, i)$} This function clips weight values for given $\gamma_c$. $TH_c$ is calculated by multiplying $\gamma_c$ and the maximum absolute value of weights. 

\vspace{0.1in}
\noindent
{\bf WeightRound$(w_c, s, \gamma_n, \gamma_s, i)$} This function performs weight rounding with given hyper-parameters. For more details on the rounding equation, please refer to the paper, equation (2) for the 1st-order rounding and equation (4) for the 2nd-order rounding.  

\vspace{0.1in}
\noindent
{\bf ActQuant$(i)$ } This function quantizes the activation values. Firstly, the function finds $\gamma_c$ to perform activation clipping through ActClip function. When the activation clipping is done, the function finds $\gamma_n$ and $\gamma_s$ for quantization through Bayesian Optimization. With the determined hyper-parameters, the function performs activation quantization. 

\vspace{0.1in}
\noindent
{\bf ActClip$(\gamma_c,i)$} This function clips activation values for given $\gamma_c$. $TH_c$ is calculated by multiplying $\gamma_c$ and the maximum absolute value of activation. 

\vspace{0.1in}
\noindent
{\bf ActRound$(\gamma_n, \gamma_s, i)$} This function performs activation rounding with given hyper-parameters. For more details on the rounding equation, please refer to the paper, equation (2) for the 1st-order rounding and equation (4) for the 2nd-order rounding.  

\vspace{0.1in}
\noindent
{\bf BiasCorr$(i)$} This function performs bias correction. The function evaluates twice: with and without the bias correction. Finally, the function returns a better result.  

\SetKwComment{Comment}{$\triangleright$\ }{}
\SetCommentSty{textnormal}
\SetKw{KwBy}{by}

\begin{algorithm*}[h]
  \caption{Overall algorithm of Q-Rater}
  \begin{multicols}{2}
  \DontPrintSemicolon
  
  \begin{tabularx}{\textwidth}{ll}
  $\bm{M}$ & A target model \\
  $\bm{M}.w^i$ &  A weight of the $i$\textsuperscript{th} layer \\
  $\bm{M}.x^i_j$  &  An input of the $i$\textsuperscript{th} layer for the $j$\textsuperscript{th} batch \\
  $N_{layer}$ &  The number of layers in $\bm{M}$ \\ 
  $f_{eval}$ &  An evaluation function with quant. parameters \\ 
  $\bm{S_{x}}$ &  Scaling factors for each activation \\ 
  $\bm{\Gamma_{n}, \Gamma_{s}}$  &  $\gamma_n$ and $\gamma_s$ for each activation \\
  $\bm{BC_{x}}$ & An indicating vector for bias-correction \\
  $q$ & The number of quantization bits \\
  \texttt{BO} & Bayesian optimizer \\
  \texttt{.probe} & registers a given point \\
  \texttt{.run} & optimizes a function within given parameter ranges \\
  \end{tabularx}
  \;
  \SetKwFunction{FCc}{Run}
  \SetKwProg{Fn}{Func}{:}{\KwRet}
  \Fn{\FCc{}}{
   \For{$i \gets 1$ \KwTo $N_{layer}$}{
        \texttt{WeightQuant($i$)} \;
        \texttt{ActQuant($i$)} \;
        \texttt{BiasCorr($i$)} \;
    }
}
\;

\SetKwFunction{FCc}{WeightQuant}
\Fn{\FCc{$i$}}{
        $w_{org} \gets \bm{M}.w^i$ \;
        \For{$\gamma_c \gets 0.1$ \KwTo $1.0$ \KwBy $0.1$}{
	        $acc \gets $\texttt{WeightClip}$(w_{org}, \gamma_c, i)$ \;
	        $\texttt{BO.probe}(\gamma_c, acc)$ \;
	    }
	    $\gamma_c^{*} \gets \texttt{BO.run}(\texttt{WeightClip}, \gamma_c=(0, 1.0)$) \;
        $Th_c \gets \gamma_c^{*} \times max(| w_{org} |)$ \;
        $w_c \gets max(min(w_{org}, Th_c), -Th_c)$ \;
        $s \gets Th_c / (2^{q-1}-1)$ \;
        \;
        \For{$\gamma_n \gets -1.0$ \KwTo $1.0$ \KwBy $0.1$}{
            \For{$\gamma_s \gets 0.0$ \KwTo $1.0$ \KwBy $0.25$}{
                $acc \gets $\texttt{WeightRound}$(w_c, s, \gamma_n, \gamma_s, i)$ \;
                $\texttt{BO.probe}((\gamma_n, \gamma_s), acc)$ \;
	        }
	    }
	    $\gamma_n^{*}, \gamma_s^{*} \gets$ \texttt{BO.run}$($\texttt{WeightRound} $,$\;
	    \Indp\Indp\Indp $\gamma_n=(-1.0, 1.0), \gamma_s=(0, 1.0))$ \;
        \Indm\Indm\Indm $\bm{M}.w^{i} \gets s \cdot \left \lfloor \frac{w_c}{s} + 0.5 + f_r(w_c, \gamma_n^{*}, \gamma_s^{*} ) \right \rfloor$ \;
}
\;

\SetKwFunction{FCc}{WeightClip}
\Fn{\FCc{$w$, $\gamma_c$, $i$}}{
    $Th_c \gets \gamma_c \times max(|w|)$ \;
    $M.w_{i} \gets max(min(w, Th_c), -Th_c)$ \;
    \KwRet $f_{eval}(\bm{M}, \bm{S_{x}}, \bm{\Gamma_{n}}, \bm{\Gamma_{s}}, \bm{BC_{x}})$ \;
    
}
\;
\SetKwFunction{FCc}{WeightRound}
\Fn{\FCc{$w$, $s$, $\gamma_n$, $\gamma_s$, $i$}}{
    $w' \gets s \cdot \left \lfloor \frac{w}{s} + 0.5 + f_r(w, \gamma_n, \gamma_s ) \right \rfloor$  \Comment*[l]{Eq. (4) 
    }
    $M.w_{i} \gets w'$ \;
    \KwRet $f_{eval}(\bm{M}, \bm{S_{x}}, \bm{\Gamma_{n}}, \bm{\Gamma_{s}}, \bm{BC_{x}})$ \;
}
\;
\;
\SetKwFunction{FCc}{ActQuant}
\Fn{\FCc{$i$}}{
        \For{$\gamma_c \gets 0.1$ \KwTo $1.0$ \KwBy $0.1$}{
            $acc \gets \texttt{ActClip}(\gamma_c, i)$ \;
            $\texttt{BO.probe}(\gamma_c, acc)$ \;
        }
        $\gamma_c^{*} \gets \texttt{BO.run}(\texttt{ActClip}, \gamma_c=(0, 1.0)$) \;
        $Th_c = (\sum_{j=1}^{n_{batch}} \gamma_c^{*} \times max|\bm{M}.x^{i}_{j}|)/ n_{batch}$ \;
        $S_{x}^{i} = Th_c / (2^{q-1}-1)$ \;
        \;
        \For{$\gamma_n \gets -1.0$ \KwTo $1.0$ \KwBy $0.1$}{
            \For{$\gamma_s \gets 0.0$ \KwTo $1.0$ \KwBy $0.25$}{
                $acc \gets \texttt{ActRound}(\gamma_n, \gamma_s, i)$ \;
	            \texttt{BO.probe}$((\gamma_n, \gamma_s), acc)$ \;
	       }
	   }
	   $\gamma_n^{*}, \gamma_s^{*} \gets$ \texttt{BO.run}$($\texttt{ActRound} $,$\;
	   \Indp\Indp\Indp $\gamma_n=(-1.0, 1.0), \gamma_s=(0, 1.0))$ \;
	   \Indm\Indm\Indm $\Gamma_n^{i}, \Gamma_s^{i} \gets \gamma_n^{*}, \gamma_s^{*}$ \;
}
\;
\;
\SetKwFunction{FCc}{ActClip}
\Fn{\FCc{$\gamma_c$, $i$}}{
    $Th_c = (\sum_{j=1}^{n_{batch}} \gamma_c \times max|\bm{M}.x^{i}_{j}|)/ n_{batch}$ \;
    $S_{x}^{i} = Th_c / (2^{q-1}-1)$ \;
    \KwRet $f_{eval}(\bm{M}, \bm{S_{x}}, \bm{\Gamma_{n}}, \bm{\Gamma_{s}}, \bm{BC_{x}})$ \;
}
\;
\;
*Activation rounding is performed dynamically \;
within evaluation function.\;
\SetKwFunction{FCc}{ActRound}
\Fn{\FCc{$\gamma_n$, $\gamma_s$, $i$}}{
    $\Gamma_n^{i}, \Gamma_s^{i} \gets \gamma_n, \gamma_s$ \;
    \KwRet $f_{eval}(\bm{M}, \bm{S_{x}}, \bm{\Gamma_{n}}, \bm{\Gamma_{s}}, \bm{BC_{x}})$ \;
}
\;
\;

\SetKwFunction{FCc}{BiasCorr}
\Fn{\FCc{$i$}}{
    $acc^{off} \gets f_{eval}(\bm{M}, \bm{S_{x}}, \bm{\Gamma_{n}}, \bm{\Gamma_{s}}, \bm{BC_{x}})$ \;
    $BC_{x}^{i} = TRUE$ \;
    $acc^{on} \gets f_{eval}(\bm{M}, \bm{S_{x}}, \bm{\Gamma_{n}}, \bm{\Gamma_{s}}, \bm{BC_{x}})$ \;
    \If{$acc^{off} > acc^{on}$}{
        $BC_{x}^{i} = FALSE$ \;
    }
}
\;

\end{multicols}
\label{alg:main-alg}
\end{algorithm*}

\end{document}

%% file: math_commands.tex
\usepackage{amsmath,amsfonts,bm}

\def\eqref#1{equation~\ref{#1}}

\def\1{\bm{1}}

\def\vh{{\bm{h}}}

\def\vx{{\bm{x}}}
\def\vy{{\bm{y}}}

\def\mW{{\bm{W}}}
\def\mX{{\bm{X}}}

\DeclareMathAlphabet{\mathsfit}{\encodingdefault}{\sfdefault}{m}{sl}
\SetMathAlphabet{\mathsfit}{bold}{\encodingdefault}{\sfdefault}{bx}{n}

\newcommand{\E}{\mathbb{E}}

\DeclareMathOperator{\sign}{sign}